\documentclass[10pt,twocolumn,letterpaper]{article}

\usepackage{iccv}              

\usepackage{pifont}
\usepackage{times}
\usepackage{microtype}
\usepackage{epsfig}
\usepackage{caption}
\usepackage{longtable} 
\usepackage{float}
\usepackage{placeins}
\usepackage{color, colortbl}
\usepackage{stfloats}
\usepackage{enumitem}
\usepackage{xstring}
\usepackage{multirow, makecell}
\usepackage{xspace}
\usepackage{url}
\usepackage{subcaption}
\usepackage{xcolor}
\usepackage{algorithm}
\usepackage{algpseudocode}
\usepackage[hang,flushmargin]{footmisc}
\usepackage{tabularx}
\usepackage{threeparttable}
\usepackage{amsmath}
\usepackage{amssymb}
\usepackage{xspace}

\makeatletter
\algdef{SE}[FOR]{ForEach}{EndForEach}[1]{%
  \algorithmicfor\ #1\ \algorithmicdo
}{%
  \algorithmicend\ \algorithmicfor
}
\makeatother

\usepackage{graphicx}
\usepackage{bm}
\usepackage{booktabs}
\usepackage{bigstrut}
\usepackage{bbding}
\usepackage[misc]{ifsym}
\newcommand{\xmark}{\ding{55}} 
\newcommand{\cmark}{\ding{51}}

\definecolor{iccvblue}{rgb}{0.21,0.49,0.74}
\usepackage[pagebackref,breaklinks,colorlinks,allcolors=iccvblue]{hyperref}

\title{MDK12-Bench: A Multi-Discipline Benchmark for \\ Evaluating Reasoning in Multimodal Large Language Models}

\author{
\hspace{-6mm}
    Pengfei Zhou\textsuperscript{1\thanks{Equal contribution $^\dagger$Corresponding author}},
    Fanrui Zhang\textsuperscript{2,3$^*$},
    Xiaopeng Peng\textsuperscript{4$^*$},
    Zhaopan Xu\textsuperscript{5,1},
    Jiaxin Ai\textsuperscript{6,2},
    Yansheng Qiu\textsuperscript{6,1}, \\
\hspace{-6mm}    Chuanhao Li\textsuperscript{1}, 
    Zhen Li\textsuperscript{1},
    Ming Li\textsuperscript{1},
    Yukang Feng\textsuperscript{2},
    Jianwen Sun\textsuperscript{2},
    Haoquan Zhang\textsuperscript{1},
    Zizhen Li\textsuperscript{2},
    Xiaofeng\\ \hspace{-6mm} Mao\textsuperscript{1},
    Wangbo Zhao\textsuperscript{8},
    Kai Wang\textsuperscript{8},
    Xiaojun Chang\textsuperscript{3,7},
    Wenqi Shao\textsuperscript{1},
    Yang You\textsuperscript{8$^\dagger$},
    Kaipeng Zhang\textsuperscript{1,2$^\dagger$} \\
\hspace{-6mm}    \textsuperscript{1}Shanghai AI Laboratory
    \textsuperscript{2}Shanghai Innovation Institute
    \textsuperscript{3}USTC
    \textsuperscript{4}RIT
    \textsuperscript{5}HIT
    \textsuperscript{6}WHU
    \textsuperscript{7}MBZUAI 
    \textsuperscript{8}NUS
}

\begin{document}
\maketitle


\begin{abstract}
Multimodal reasoning, which integrates language and visual cues into problem solving and decision making, is a fundamental aspect of human intelligence and a crucial step toward artificial general intelligence. However, the evaluation of multimodal reasoning capabilities in Multimodal Large Language Models (MLLMs) remains inadequate. Most existing reasoning benchmarks are constrained by limited data size, narrow domain coverage, and unstructured knowledge distribution. To close these gaps, we introduce MDK12-Bench, a multi-disciplinary benchmark assessing the reasoning capabilities of MLLMs via real-world K-12 examinations. Spanning six disciplines (math, physics, chemistry, biology, geography, and information science), our benchmark comprises 140K reasoning instances across diverse difficulty levels from primary school to 12th grade. It features 6,827 instance-level knowledge point annotations based on a well-organized knowledge structure, detailed answer explanations, difficulty labels and cross-year partitions, providing a robust platform for comprehensive evaluation. Additionally, we present a novel dynamic evaluation framework to mitigate data contamination issues by bootstrapping question forms, question types, and image styles during evaluation. Extensive experiment on MDK12-Bench reveals the significant limitation of current MLLMs in multimodal reasoning. The findings on our benchmark provide insights into the development of the next-generation models. Our data and codes are available at \href{https://github.com/LanceZPF/MDK12}{https://github.com/LanceZPF/MDK12}.

\end{abstract}    
\section{Introduction}
\label{sec:intro}

Reasoning is fundamental to human intelligence, enabling logical, rational, and deliberate thought, inference, and deduction beyond prior knowledge \cite{sternberg1982reasoning, lohman2011intelligence}. By integrating vision, language, and symbols, multimodal reasoning improves problem-solving and decision-making abilities based on diverse information sources. Replicating sophisticated and context-aware multimodal reasoning capabilities is crucial to achieving Artificial General Intelligence (AGI)~\cite{morris2023levels}. 

With the rapid advancement of Multimodal Large Language Models (MLLMs)~\cite{radford2021learning, li2022blip, liu2023llava, alayrac2022flamingo}, reliable benchmarks are needed to assess their real-world reasoning capabilities. Existing multimodal benchmarks focus mainly on basic understanding and low-order reasoning tasks~\cite{kilmllm, rajabi2024gsr,hudson2019gqa}, such as numerics \cite{tanaka2023slidevqa}, common sense \cite{yingmmt,liu2024mmbench} and image quality judgment \cite{yang2024seed}. Unlike low-order reasoning, which relies on common knowledge, high-order reasoning requires step-by-step thinking and systematic analysis capabilities. While most existing benchmarks are restricted to single isolated disciplines (e.g., mathematics \cite{baral24drawedumath,lumathvista, wang2025measuring} and medicine \cite{xiacares, sun2024pathmmu}) or common knowledge \cite{hao2025can, yue2024mmmu}, the evaluation of high-order reasoning has not been fully explored. 

While several early attempts have been made to evaluate the complex reasoning performance of MLLMs \cite{huang2024survey, li2024survey}, the previous benchmarks still have limitations in data scope, data size, data granularity, or systematic knowledge structuring. As a result, these benchmarks lack the breadth and depth to challenge the reasoning capabilities of MLLMs in complicated real-world reasoning tasks. In addition, due to the absence of fine-grained key-point annotations and structured knowledge, these benchmarks fail to trace how MLLMs fail in solving certain problems.

To address these challenges, we introduce MDK12-Bench, a multi-disciplinary benchmark assessing reasoning capabilities of MLLMs at the K-12 level (defined for simplicity as Grades 1 to 12 excluding kindergarten). Spanning from Grade 1 to Grade 12, K-12 education is deeply interwoven with disciplinary examinations for testing knowledge comprehension and high-order thinking skills \cite{liu2024k, chen2024systematic}. In contrast to higher education, where individuals receive in-depth knowledge for professional development through self-guided learning, K-12 offers a broad spectrum of subjects that are more structured, well-defined and interconnected. These characteristics make the K-12 domains an ideal testbed for systematically evaluating the knowledge coverage, reasoning, and problem-solving abilities of MLLMs.

\begin{table*}[!htbp]
  \centering
  \small
   \begin{tabularx}{\textwidth}{@{}
    >{\raggedright\arraybackslash}p{2.5cm} 
    >{\centering\arraybackslash}p{1.6cm} 
    >{\centering\arraybackslash}p{1.2cm} 
    >{\centering\arraybackslash}p{1.0cm} 
    >{\centering\arraybackslash}p{3.2cm} 
    >{\centering\arraybackslash}p{1.1cm} 
    >{\centering\arraybackslash}p{1.3cm}  
    >{\centering\arraybackslash}p{1.1cm}  
    >{\centering\arraybackslash}p{1.1cm}  
    @{}}
    \toprule
    \multicolumn{1}{c}{\multirow{2}[4]{*}{Benchmarks}}  & 
    \multicolumn{4}{c}{Data Coverage} & 
    \multirow{2}[4]{*}{Modality} & 
    \multirow{2}[4]{*}{\parbox{1.2cm}{Explanation \\ Annotation}} & 
    \multirow{2}[4]{*}{\parbox{1.2cm}{\centering Structured \\ Knowlege}} & 
    \multirow{2}[4]{*}{\parbox{1.2cm}{\centering Dynamic\\Evaluation}} \\
    \cmidrule{2-5} & 
    \multicolumn{1}{c}{Level}& \#Instances  & \#Images  & Question Type   \\
    \midrule
    MMBench \cite{liu2024mmbench}  
         &Low-order & 3.2K &- & MC & I+T &\xmark & \xmark  & \xmark \\
    MMIU \cite{mengmmiu}  
         &Low-order & 11.6K &77K & MC, Open & I+T &\xmark & \xmark  & \xmark \\
    MMT-Bench\cite{yingmmt}  
         &Low-order  & 31.2K &31.2K & MC & I+T &\xmark& \xmark  & \xmark \\
    EMMA \cite{hao2025can}  
         &College  & 2.7K &3K& MC, Open & I+T &\xmark& \xmark  & \xmark \\
    MMMU \cite{yue2024mmmu}  
         &College  & 11.5K &12.3K  &  MC, Open & I+T & \xmark& \xmark  & \xmark \\
    DrawEduMath \cite{baral24drawedumath}  
         &K12-Math  & 44K &2.3K  &  Open & I+T &\cmark & \xmark  & \xmark \\
    \midrule
    MDK12-Bench 
         & K-12 & \textbf{141.3K} & \textbf{105.2K}& \textbf{MC, Fill, T/F, Open}& \textbf{T, I+T} & \cmark & \cmark & \cmark \\
    \bottomrule
    \end{tabularx}
    \caption{\textbf{Comparison between our MDK12-Bench and existing multimodal reaonsing benchmarks.} MDK12-Bench includes more comprehensive data and question coverage. The systematic knowledge structuring and dynamic test-time augmentation also provide more reliable and fair evaluation of MLLMs. T: Text; I: Image. MC: multiple-choice; Open: open-ended; Fill: fill-in-the-blank; T/F: true or false; Low-order reasoning: commonsense, image quality judgement, relation, attribute reasonings, etc.}
  \label{tab:dataset_comparison}%
\end{table*}%

\begin{figure*}[t]
	\centering
	\includegraphics[trim=0cm 0.3cm 0 0, width=0.97\textwidth]{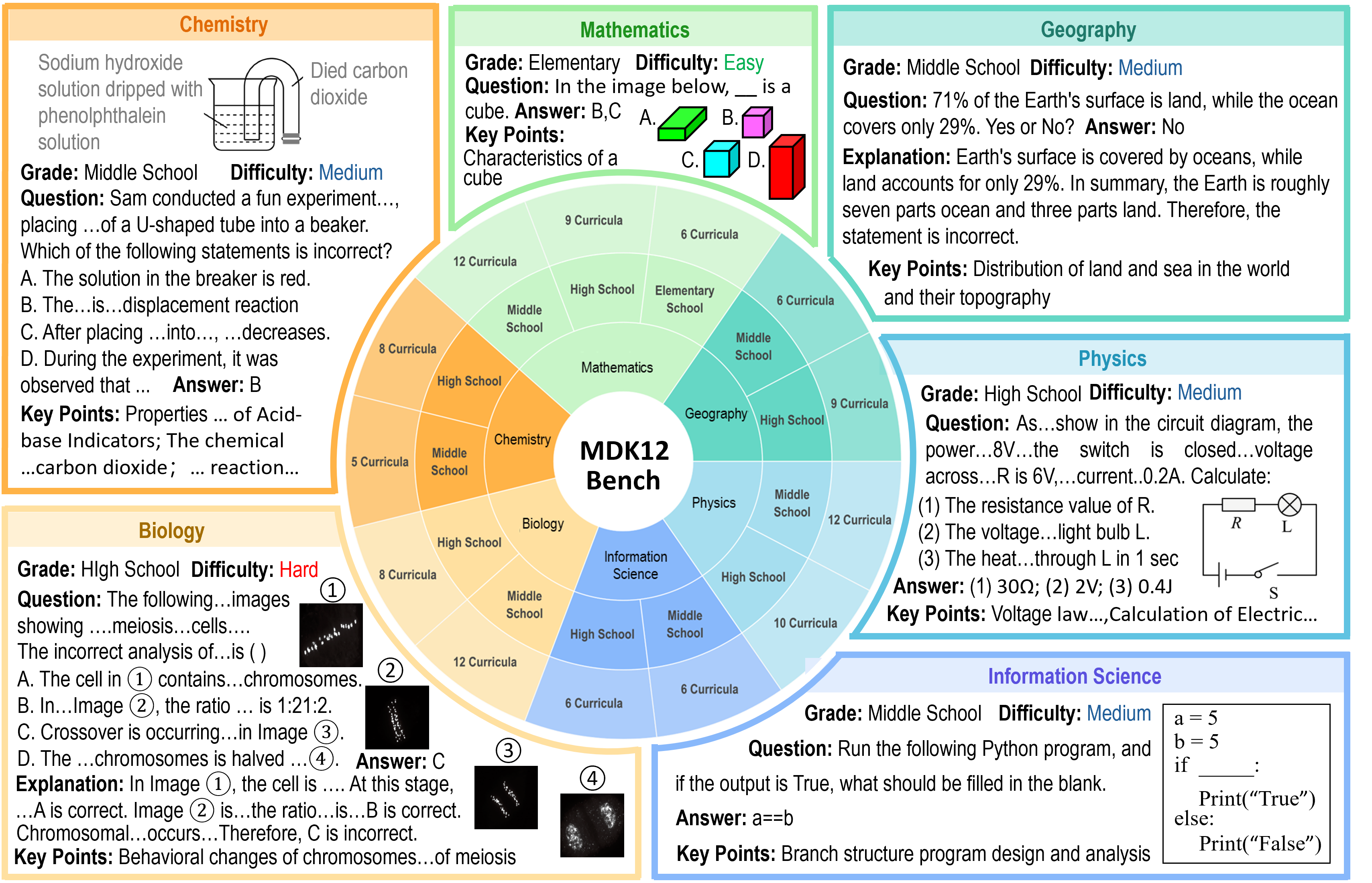}
  \caption{\textbf{Overview of MDK12-Bench.} It comprises 140K instances and spans 6 disciplines in K-12 education. The knowledge system of our bench is structured into six fine-grained levels: discipline, grade, curriculum, topic, meta-knowledge, and key knowledge points, where the three rings showcase the first three levels. Examples illustrate the representative grades (elementary, middle, and high schools), difficulty levels (easy, medium, and high), questions and answers, and key knowledge points of each discipline. The diverse question forms (single- and multiple-choice, open-ended question, fill-in-blank, true-or-false) and detailed answer explanations are also demonstrated.}
	\label{fig:data}
\end{figure*}

As shown in Table \ref{tab:dataset_comparison} and Fig. \ref{fig:data}, our MDK12-Bench consists of 141.3K reasoning questions and spans 6 reasoning-oriented K-12 disciplines: Mathematics, Physics, Chemistry, Biology, Geography, and Information Science. For each discipline, we provide fine-grained annotations including difficulty levels, instance-level key knowledge points, and detailed answer explanations. With cross-year partitions, the MDK12-Bench allows various breakdown analyses, cross-validations, and dynamic updates. Moreover, the instance-level key knowledge point annotation is linked with our constructed knowledge tree, enabling deeper model performance analysis at each knowledge level. Compared to existing benchmarks, our MDK12-Bench potentially provides a more systematic evaluation of MLLMs' multimodal reasoning capabilities in real-world academic tasks.

In addition, current reasoning model evaluations are typically static, which can be biased due to data contamination, i.e., test items appearing in the MLLM’s large-scale training data. To address this, we introduce a novel dynamic evaluation framework that automatically transforms both the textual and visual parts of questions via different bootstrapping strategies, including word substitution, paraphrasing and question type, permuting for textual bootstrapping and image expansion, color shift, and style transfer for visual bootstrapping. Based on MDK12-Bench’s diverse content, this dynamic framework offers a more robust and fair platform for evaluating high-order multimodal reasoning in MLLMs.

We evaluate various classic and state-of-the-art MLLMs on our MDK12-Bench using the proposed dynamic evaluation method. Extensive experiment results demonstrate that large models trained with reasoning-related data, such as Gemini2.0-flash-thinking and QVQ-72B, generally perform better than the smaller common models. It has also been proven that model performance under the dynamic evaluation setting can face more challenges than the original benchmarks. 
Our contributions are summarized as follows: 
\begin{itemize}
    \item \textbf{A Comprehensive Multi-Discipline Benchmark.} We present MDK12-Bench, a systematically curated, large-scale K12-based benchmark supporting the comprehensive evaluation of the reasoning capability of MLLMs.
    \item \textbf{A Dynamic Evaluation Framework.} A practical framework mitigating data contamination and providing flexible multimodal data transformation to challenge MLLMs with bootstrapped unseen data.
    \item \textbf{A Comprehensive Leaderboard.} We provide a detailed analysis of current MLLMs in our leaderboard. Our studies suggest that both our subsets and dynamic evaluation benchmark challenge the reasoning capabilities of current MLLMs, showing that our work can support a robust platform for evaluating reasoning-oriented models and challenging future artificial general intelligence attempts.
\end{itemize}

\section{Related Works}
\label{sec:formatting}

\noindent\textbf{MLLM Reasoning.}
Based on the rapid advances of Large Language Models (LLMs)~\cite{openai2023gpt,team2023gemini,touvron2023llama}, multimodal LLMs (MLLMs) have emerged to address complex multimodal tasks, including interleaved image-text generation \cite{zhou2024gate,touvron2023llama,gpt4v,team2023gemini,claud3,team2023gemini}. Recent studies have also proposed domain-specific MLLMs utilizing multimodal pretraining, vision instruction tuning, and reinforcement learning, such as Math-LLaVA \cite{shi2024math} and MultiMath \cite{peng2024multimath} for mathematical tasks, and Med-Flamingo \cite{moor2023med}, LLaVA-Med \cite{liu2023llava}, and Med-MoE \cite{jiang2024med} in the biomedical domain. Furthermore, methods like Chain-of-Thought (CoT) prompting \cite{wei2022chain}, iterative bootstrapping techniques and reinforce learning~\cite{zhong2024dpo}, such as STaR \cite{zelikman2024star}, Quiet-STaR \cite{zelikman2024quiet} and DeepSeek-E1~\cite{guo2025deepseek}, have further improved interpretability and response quality.

\noindent\textbf{MLLM Evaluation.}
With the rapid advancement of MLLMs, various benchmarks have been proposed to assess their performance~\cite{saikh2022scienceqa,lumathvista,zhang2024mathverse,mengmmiu}. However, most benchmarks focus primarily on fundamental perceptual skills, and lack the evaluation of expert-level domain knowledge and deep reasoning, or include reasoning only in limited contexts. For instance, MathVerse \cite{zhang2024mathverse} emphasizes visual mathematical comprehension, while MMBench examines basic visual understanding and cross-modal fusion~\cite{liu2024mmbench}. 
Recently, more comprehensive evaluations have emerged. For instance, MMMU \cite{yue2024mmmu} proposed a large-scale "expert AI" challenge across domains like medicine and law. GSM-8K \cite{cobbe2021training} emphasized logical consistency in mathematical reasoning. EXAMS-V \cite{das2024exams} extended multilingual, multimodal exam-style assessments. Despite these advances, existing benchmarks still face limitations in data size, knowledge system completeness, and domain diversity, as they often focus on mathematics or a narrow set of specialized knowledge. Our benchmark aims to overcome these limitations by evaluating complex reasoning and multi-domain proficiency.

\noindent\textbf{Dyanmic Evaluation.}
Growing doubts about the “true capabilities” of LLMs in public benchmarks often attribute “spurious performances” to data contamination. To keep pace with evolving model capabilities and mitigate contamination, researchers have turned to dynamic or adaptive evaluations. Zhu et al. \cite{zhu2024dynamic} introduced a meta-probing agent that continually adjusts test content and difficulty during fine-tuning or domain adaptation, while Yang et al. \cite{yang2024dynamic} dynamically modifies visual and textual context to verify the influence of data contamination.
These dynamic methods indicate a promising future for benchmarks, which must evolve as models expand their training corpora. More robust protocols are needed to distinguish genuine reasoning improvements from mere pattern matching. Therefore, we propose a dynamic testing framework for MDK12-Bench to counter data leakage and maintain sustained benchmark integrity.

\section{MKD12-Benchmark}
\begin{figure*}[t]
	\centering
	\includegraphics[trim=0.5 0.5cm 0 0, width=0.98\textwidth]{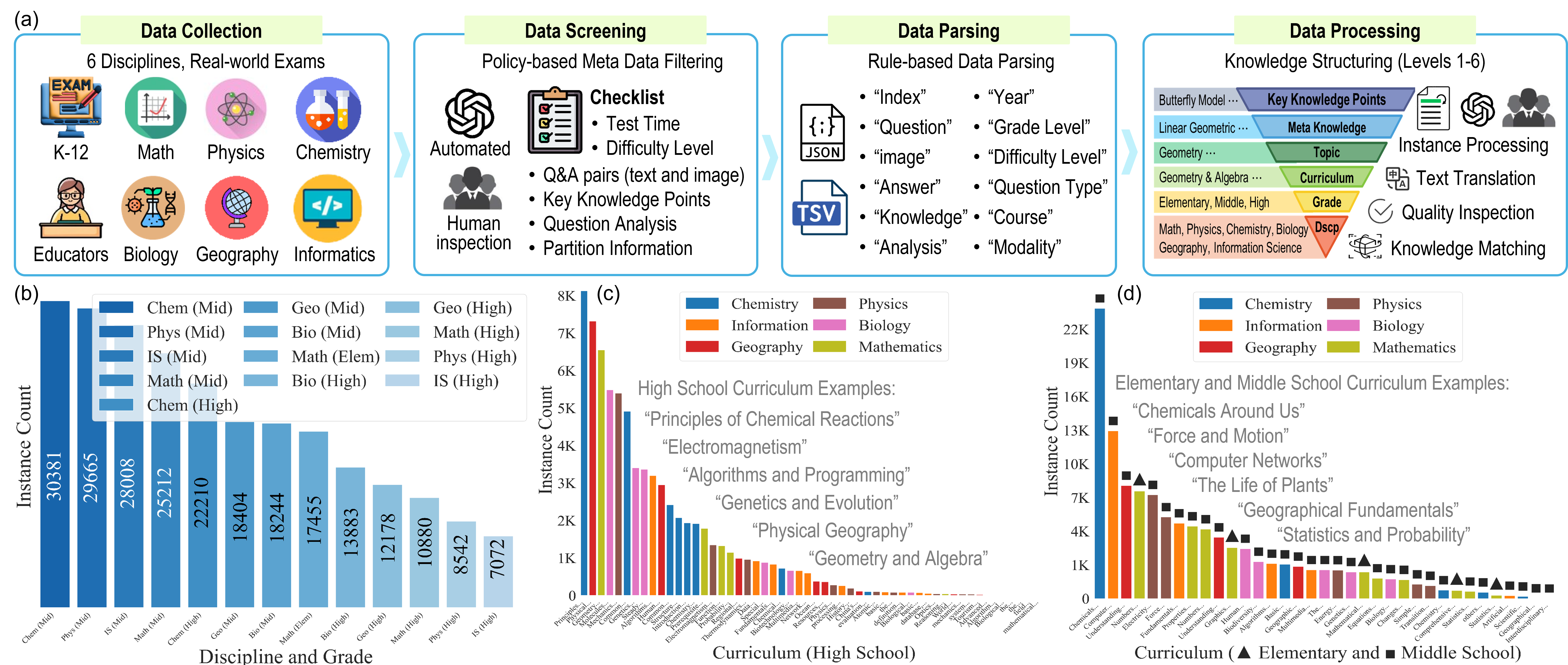}
   \caption{\textbf{Data curation and statistics of our MDK12-Bench.} (a) The data curation pipeline consists of four stages: data collection, screening, parsing, and processing. The knowledge in our benchmark is structured into six interconnected levels: Level 1 - discipline, Level 2 - grade, Level 3 - curriculum, Level 4 - topics, Level 5 - meta-knowledge, and Level 6 - key knowledge point. Statistics of knowledge coverage of our bench is illustrated in terms of the number of instance occurrences at (b) discipline and grade levels; (c) high-school curriculum level; and (d) elementary- and middle-school curriculum level. Examples of curriculum-level knowledge points are also demonstrated. }
\label{fig:data_curation}
\end{figure*}

To address the scarcity of high-quality multimodal academic reasoning benchmarks, We created MKD12-Bench in around two months with the participation of over 20 researchers and several K-12 educators. The data curation of our benchmark involves four stages, as shown in Fig. \ref{fig:data_curation}(a).

\noindent\textbf{Data Collection.}
The data collection involves an extensive search of online open-source exam paper repositories. The team established and refined an efficient acquisition workflow to ensure both reliability and broad coverage of the collected questions, laying a robust foundation for subsequent screening and parsing. We accumulated large-scale question sets from multiple grade levels, regions, and multimodal formats to form the metadata source for MDK12-Bench.


\noindent\textbf{Data Screening.}
After the initial collection, we utilized a combination of GPT-4o-based automated review and human inspection, based on a predefined checklist, to filter the invalid metadata. We removed questions containing low-quality images or without specific knowledge points. Additionally, we preserved the question analysis section in each instance to support chain-of-thought studies, along with partition information covering exams from 2016 to 2025, thus allowing examination of potential knowledge evolution over time. Each question instance was assigned a difficulty level, facilitating dynamic evaluations and enabling model performance analysis under varying complexity levels.

\noindent\textbf{Data Parsing.}
Following the screening stage, we performed rule-based parsing to transform each question into a structured format. This process extracted fields such as Year, Question, Grade Level, Image, Difficulty Level, Answer, Question Type, Knowledge, Course, Analysis, and Modality, all of which were systematically stored. By aligning questions, solutions, and annotations in a standardized structure, the resulting dataset significantly enhances retrieval efficiency, allowing models or analysis tools to readily access specific subjects, difficulty levels, or knowledge points.

\noindent\textbf{Data Processing.}
Once the data was parsed, we carried out additional post-processing steps to ensure linguistic and formatting consistency. Leveraging the GPT-4o API, we translated all Chinese text into English, followed by meticulous domain-expert reviews to verify technical accuracy. For images containing Chinese text, we utilized an image translation tool and performed manual checks. Furthermore, we built a comprehensive knowledge structure encompassing discipline, grade/year, curriculum, chapter/unit, lesson/topic, and key knowledge points, linking each translated question to this framework to enrich its academic context. 

\begin{table}[htbp]
\centering
\small
\renewcommand{\arraystretch}{0.7}
\caption{Key Statistics of MDK12-Bench}
\label{tab:stats}
\begin{tabular}{ll}
\toprule
\multicolumn{2}{c}{\textbf{Overall Statistics}} \\
\midrule
Total instances         & 141,320 \\
Text-only instances     & 77,857  \\
Multimodal instances  & 63,463   \\
Total images            & 105,218 \\
Exam years coverage & 10  \\
\midrule\midrule
\multicolumn{2}{c}{\textbf{Knowledge Structure}} \\
\midrule
Knowledge levels & 6 \\
Total knowledge points & 6,827 \\
Level 1\&2 knowledge points & 13 \\
Level 3 knowledge points & 90 \\
Level 4 knowledge points & 499 \\
Level 5\&6 knowledge points & 6,225 \\
\midrule\midrule
\multicolumn{2}{c}{\textbf{Mini-Subsets}} \\
\midrule
Total instances & 14,595 \\
Easy-level instances       & 4.951 \\
Medium-level instances       & 4,692 \\
Hard-level instances       & 4,952 \\
\bottomrule
\end{tabular}
\end{table}

\noindent\textbf{Data Statistics.} As shown in Table~\ref{tab:stats} and Fig.~\ref{fig:data_curation}(b), our benchmark comprises 141,320 instances, including 77,857 text-only and 63,463 multimodal instances, covering a total of 105,218 images over 12 years (2016-2025). Additionally, we define four question formats: multiple-choice (single-answer and multi-answer), fill-in-the-blank, true-or-false, and open-ended (primarily mathematical calculation). For accelerating assessment, we introduce \textbf{MDK12-Mini}, which includes three datasets. Each consists of 10\% of the data from MDK12-Bench sampled at easy, medium, and hard levels respectively. Key knowledge points are uniformly sampled to ensure that each subset includes at least one instance per key knowledge point (each instance may cover multiple knowledge points). As MLLMs' reasoning capabilities can also be challenged by text-only prompts, we include both single-modal and multimodal instances in MDK12-Bench.







\section{Dynamic MLLM Evaluation} \label{sec:dynamic_bootstrapping_eval}

\begin{figure*}[t]
	\centering
	\includegraphics[trim=0.5 0.5cm 0 0, width=0.98\textwidth]{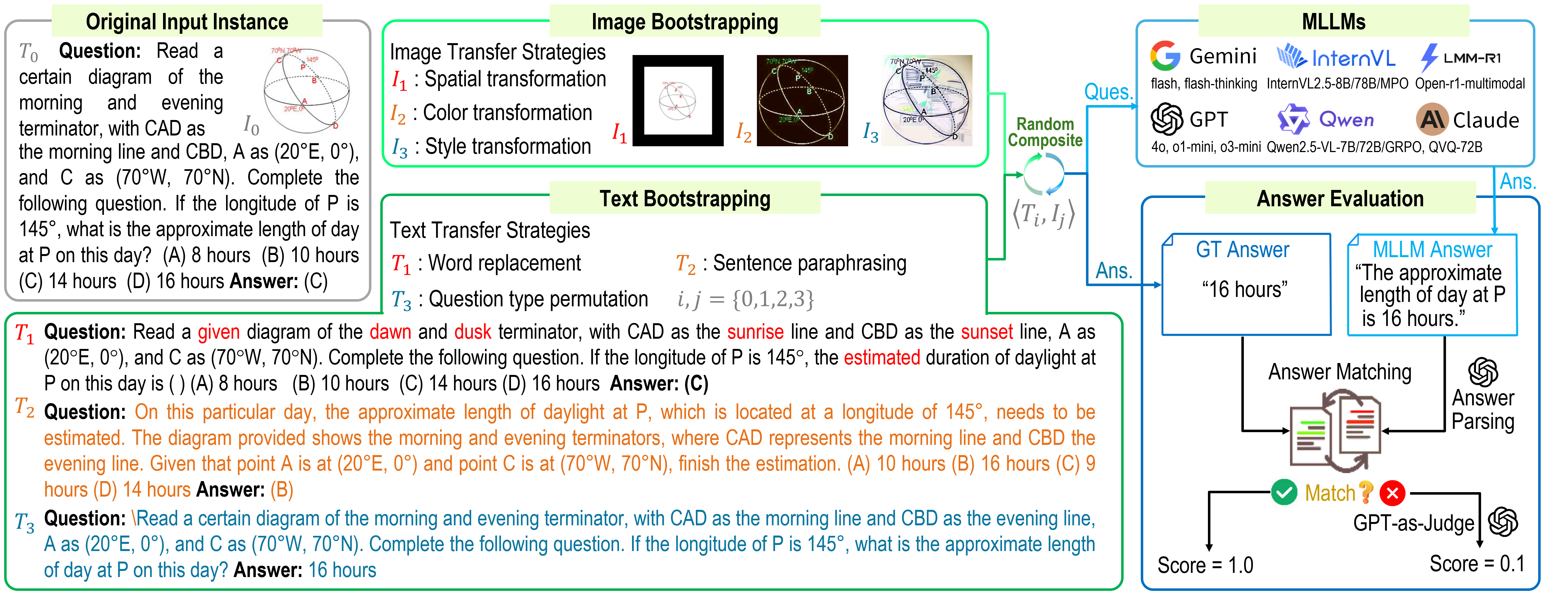}
   \caption{\textbf{The proposed dynamic MLLMs evaluation pipeline.} It includes an image and a text bootstrapping module to mitigate data contamination and a two-stage answer evaluation module comparing the model answers with ground truth.}
\label{fig:dynamic_eva}
\end{figure*}


\subsection{Bootstrapping Methods}
As illustrated in Fig.~\ref{fig:dynamic_eva}, we propose a dynamic evaluation framework that introduces controlled perturbations to texts and images during the evaluation of MLLMs by creating new test samples while preserving the accuracy of the answers. 
\paragraph{Image Bootstrapping.} Three strategies are proposed to improve image diversity without changing image semantics: 
\begin{itemize} 
\item \textbf{Spatial Transformation.} We pad the original image with colors uniformly sampled from black, white, and grey. The padding width is proportional to the image dimension along each side, with the ratios uniformly sampled in the range between 10$\%$ and 20$\%$. The image padding allows the evaluation of the model’s recognition and localization performance in varying spatial visual contexts. 
\item \textbf{Color Transformation.} In this step, the colors of the original image were inverted. Salt-and-pepper noise of random noise density is also added. This transformation assesses the model’s resilience to significant color distortions and random visual artifacts, ensuring it can still accurately identify and reason about objects. 
\item \textbf{Style Transformation.} We apply mild style transformations using the  Flux-Dev \cite{blackforestlabs_flux} model, introducing subtle style variations without significantly altering its key visual elements and semantics that relate to the question. This tests the model’s robustness to image style shifts. 
\end{itemize}

\paragraph{Textual Bootstrapping.} We introduce three methods to modify questions while preserving the answer's correctness: 
\begin{itemize} 
\item \textbf{Word Substitution.} We replace certain keywords with synonyms or contextually related expressions. This tests how well a model can maintain an accurate understanding when familiar terms are changed, thus assessing vocabulary sensitivity and semantic generalization. 
\item \textbf{Sentence Paraphrasing.} We rephrase entire sentences through variations in sentence structure, word order, or style. This checks whether a model can consistently capture the underlying meaning even when the surface form of the text is altered. 
\item \textbf{Question Type Permutation.} We convert a question from one format to another, such as turning a multiple-choice problem into a fill-in-the-blank. By changing the required style of the answer, we can see if the model retains key information under different response formats. 
\end{itemize}

Throughout the generation process, we apply a GPT-based judge to reject sampling wrong adapted instances, guaranteeing that each dynamically altered text and image still aligns with the question’s original correct answer. By this framework, we construct diverse versions of the sampled data, probing reasoning skills under varying degrees of complexity and avoiding serious data contamination.

\newcolumntype{Y}{>{\centering\arraybackslash}X}

\begin{table*}[!htbp]
  \centering
  \small
  \begin{tabularx}{0.99\textwidth}{@{}
    >{\raggedright\arraybackslash}p{2.9cm} 
    >{\raggedright\arraybackslash}p{0.75cm} 
    *{18}{Y}                         
    @{ \hspace{0.08cm} }
  @{}}
    \toprule
    \multicolumn{1}{l}{\multirow{3}{*}{\hspace{-1.5mm}\textbf{Models}}} 
      & \multirow{3}{*}{\textbf{Overall}} 
      & \multicolumn{6}{c}{\textbf{Easy}}
      & \multicolumn{6}{c}{\textbf{Medium}}
      & \multicolumn{6}{c}{\textbf{Hard}} \\
    \cmidrule(lr){3-8}\cmidrule(lr){9-14}\cmidrule(lr){15-20}
& 
& Math & Phys & Chem &Bio & Geo & IS
& Math & Phys & Chem &Bio & Geo & IS
& Math & Phys & Chem &Bio & Geo & IS \\
\midrule
Gemini2-thinking & \textbf{59.4} & \textbf{60.9}  & \textbf{56.1}  & \textbf{70.3}  & 69.8  & 59.1  & 65.3  & \textbf{52.8}  & \textbf{52.0}  & \textbf{67.0} & \textbf{68.8}  & 57.2  & 59.3 & \textbf{48.0}  & 55.0  & \textbf{62.7}  & 58.0  & 64.1  & \textbf{67.2} \\
Gemini2-flash  & 57.2 & 51.8 & 53.8 & 66.2 & 66.0 & 55.3 & 62.0  & 48.9 & 48.6 & 63.2 & 65.0 & 53.5 & 55.4 & 44.8 & 51.2 & 59.0 & 54.8 & 60.4 & 63.3  \\
Claude-3.7 & 49.8 & 54.3 & 47.1 & 59.8 & 63.3 & 50.9 & 55.0 & 44.8 & 43.7 & 56.4 & 52.9 & 48.2 & 51.1 & 38.3 & 44.0 & 49.2 & 48.6 & 45.2 & 49.9 \\
GPT-o1-mini & 53.1 & 53.0 & 53.8 & 42.3 & 55.7 & 55.2 & 63.1 & 47.6 & 44.6 & 46.9 & 55.1 & 50.9 & \textbf{64.8} & 40.9 & 54.4 & 47.5 & 52.7 & \textbf{64.7} & 64.6 \\
GPT-4o & 50.0 & 51.6 & 55.3 & 61.4 & 55.3 & 46.5 & 57.6 & 44.7 & 46.5 & 50.1 & 56.7 & 49.8 & 40.7 & 36.1 & 46.1 & 58.3 & 54.2 & 49.5 & 49.7  \\
QVQ-72B                       & 53.2 & 45.0 & 51.5 & 69.3 & 58.4 & 48.6 & 56.4 & 46.9 & 43.0 & 49.2 & 55.7 & \textbf{57.9} & 59.0 & 45.8 & \textbf{63.2} & 54.1 & \textbf{60.2} & 58.0 & 58.3 \\
Qwen2.5-VL-72B                & 51.9 & 44.7 & 48.8 & 54.9 & 63.7 & 57.9 & 64.8 & 40.2 & 43.0 & 50.8 & 57.9 & 47.9 & 56.9 & 43.0 & 45.7 & 50.4 & 53.1 & 53.0 & 64.6\\
Qwen2.5-VL-7B                 & 47.9 & 44.9 & 54.6 & 50.8 & 62.0 & 44.4 & 31.9 & 41.0 & 46.0 & 49.3 & 59.7 & 41.0 & 26.6 & 38.4 & 48.0 & 41.5 & 57.1 & 49.5 & 31.0 \\
Qwen2-VL-7B-GRPO   & 44.1 & 37.6 & 48.7 & 46.2 & 52.1 & 45.8 & 38.7 & 40.7 & 45.0 & 45.2 & 48.9 & 44.7 & 35.7 & 42.1 & 46.3 & 43.1 & 47.8 & 45.2 & 38.4 \\
Qwen2-VL-7B                 & 43.8 & 38.2 & 55.4 & 45.7 & 58.7 & 44.2 & 35.4 & 31.6 & 45.0 & 46.6 & 54.5 & 39.4 & 18.8 & 33.3 & 45.4 & 41.1 & 54.0 & 42.2 & 31.2 \\
InternVL2.5-MPO               & 51.7 & 51.3 & 51.2 & 63.1 & \textbf{80.0} & \textbf{62.4} & \textbf{69.9} & 36.4 & 38.8 & 45.5 & 52.5 & 47.4 & 44.1 & 38.2 & 42.6 & 55.9 & 41.0 & 53.9 & 55.5 \\
InternVL2.5-78B               & 48.2 
& 43.1 & 54.8 & 42.2 & 60.3 & 58.6 & 42.7 
& 38.2 & 47.2 & 43.7 & 46.1 & 57.3 & 38.9 
& 35.5 & 45.3 & 40.9 & 43.0 & 54.2 & 40.8 \\
InternVL2.5-8B                & 37.7 & 33.2 & 47.4 & 35.0 & 50.0 & 51.5 & 35.9 & 27.8 & 38.5 & 36.9 & 39.8 & 50.7 & 32.0 & 27.8 & 36.7 & 31.5 & 33.9 & 44.6 & 33.9 \\

\bottomrule
    \end{tabularx}
    \caption{\textbf{Performance of MLLMs on six disciplines (mathematics, physics, chemistry, biology, geography, and information science) across three difficulty levels.} The overall performance indicates the average accuracy across all grades, difficulty levels and disciplines.}
  \label{tab:perf_tab1}%
\end{table*}%

\begin{figure*}[t]
    \centering
\includegraphics[trim=0cm 0.5cm 0 0, width=0.98\textwidth]{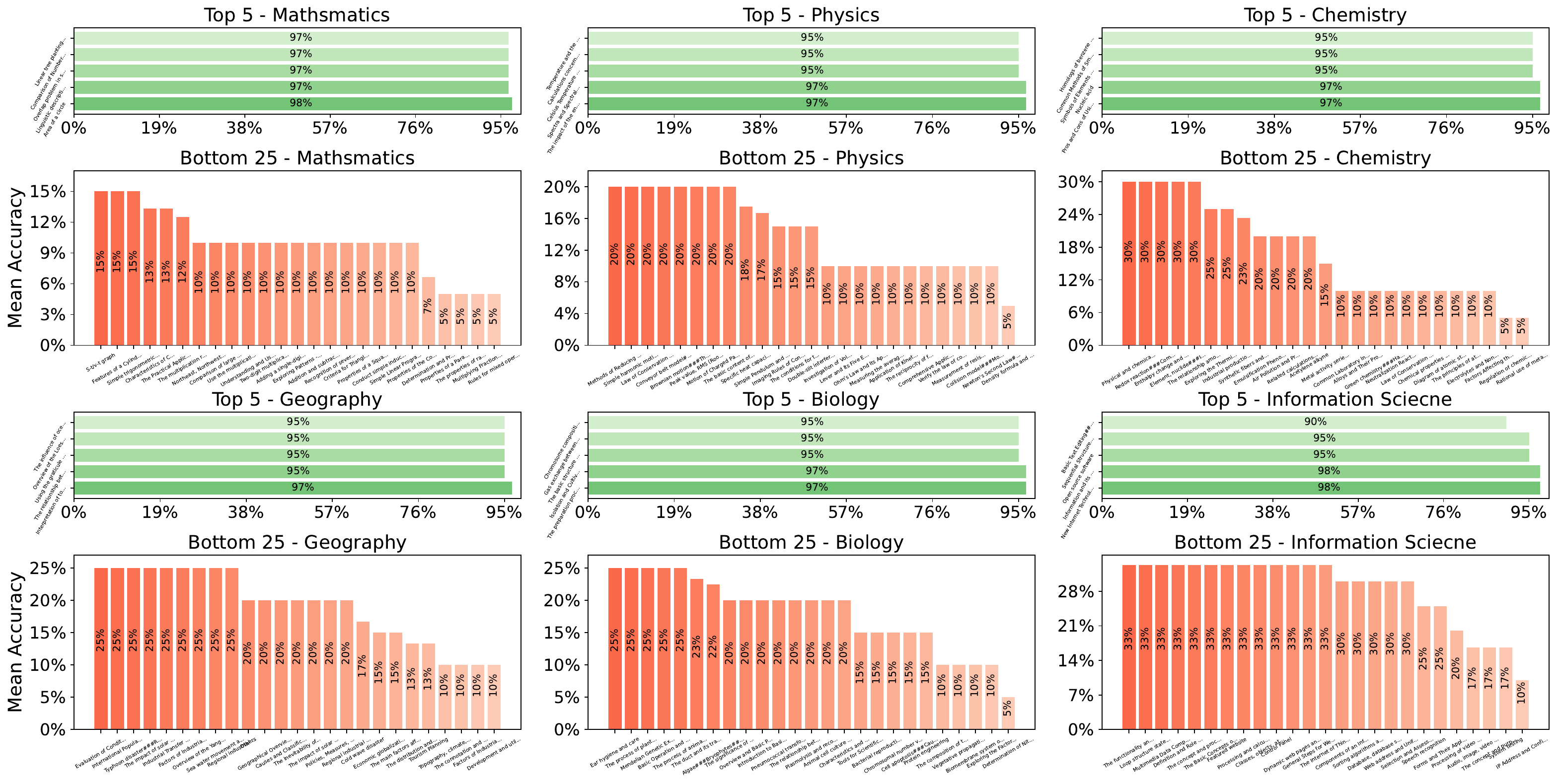}
    \caption{\textbf{Knowledge points (Level 5 - Meta Knowledge) ranked by mean accuracy of Gemini2-thinking on MDK12-Mini dataset.}}
    \label{fig:knowledge_breakdown}
\end{figure*}

\begin{figure*}[t]
    \centering
    \centering
    \includegraphics[trim=0cm 0.5cm 0 0, width=0.9\textwidth]{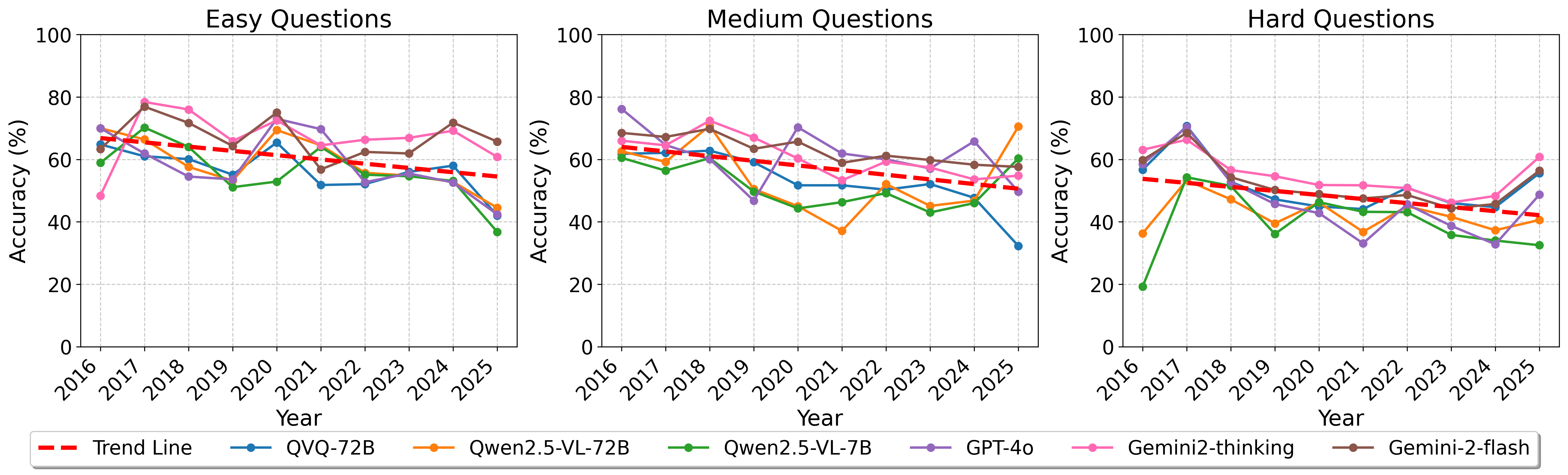}
        \caption{\textbf{Breakdown of accuracy on MDK12-Mini across different exam years.}}
    \label{fig:year_breakdown}
\end{figure*}

\subsection{Evaluation Procedures}

Our evaluation process assesses model responses through multiple steps, as illustrated in Fig.~\ref{fig:dynamic_eva}. 1) We input either the original or dynamically augmented question into the model. The model then generates a response based on textual and visual information. 2) The output is parsed using GPT as an interpreter to extract the final predicted answer from the model's response.  3) We compare the extracted answer with the ground truth. If they match exactly, the model receives full credit (a score of 1.0 for that question). If the answer does not match perfectly, we conduct a more fine-grained check with GPT and pre-defined scoring rules. If a question has multiple sub-questions or answers, we count how many elements are correct. For instance, if a fill-in-the-blank question has two blanks and the model only fills one correctly, we assign a score of 0.5.

\begin{figure}[t]
    \centering
    \includegraphics[trim=0cm 0.5cm 0 0, width=0.95\linewidth]{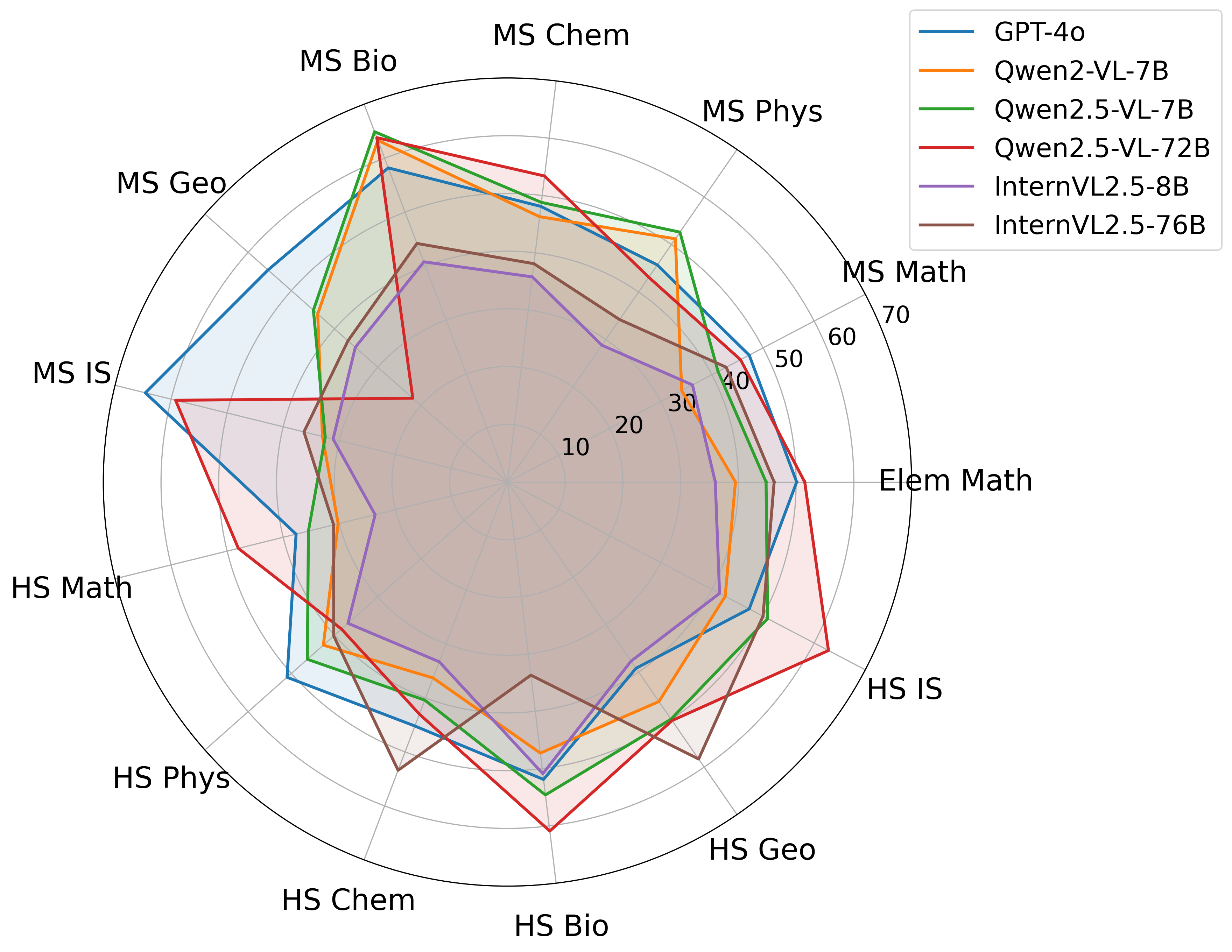}
    \caption{\textbf{Accuracy of MLLMs on full-set of MDK12-Bench.} We demonstrate the results across six disciplines (mathematics, physics, chemistry, biology, geography, and information science) and three grade levels (elementary, middle school, and high school).}
    \label{fig:full_set}
\end{figure}

\section{Experiments}
\label{sec:experiments}


\subsection{Baselines and Setup}
\label{subsec:baseline_setup}

We evaluate a set of both closed-source and open-source MLLMs, including: 1) \textbf{Closed-source MLLMs:} Gemini-2.0-flash-exp \cite{team2023gemini} (Gemini2-flash), Gemini-2.0-flash-thinking-exp \cite{team2023gemini} (Gemini2-thinking), GPT-4o \cite{openai2024gpt4o}, GPT-o1-mini \cite{openai2024gpto1mini}, GPT-o3-mini \cite{openai2025gpto3mini}, Claude-3.7-Sonnet (Claude-3.7) \cite{claud3}. 2) \textbf{Open-source MLLMs:} Qwen2.5-VL \cite{bai2025qwen2}, InternVL2.5 \cite{chen2024expandinginternvl2.5}, QVQ-72B-preview (QVQ-72B) \cite{qvq-72b-preview}, InternVL2.5-78B-MPO (InternVL2.5-MPO) \cite{wang2024mpo}, etc. 3) \textbf{Reasoning-oriented R1 variants:} Qwen2-VL-7B-GRPO-8K~\cite{openr1multimodal}. 
We mainly test the largest available models with certain reasoning capabilities, with smaller ones included for comparison if resources allow. All models are evaluated on three difficulty subsets (easy, medium, and hard). Further technical details will be discussed in the supplement.

\begin{figure}[t]
    \centering
    \includegraphics[trim=0cm 0.5cm 0 0, width=1.0\linewidth]{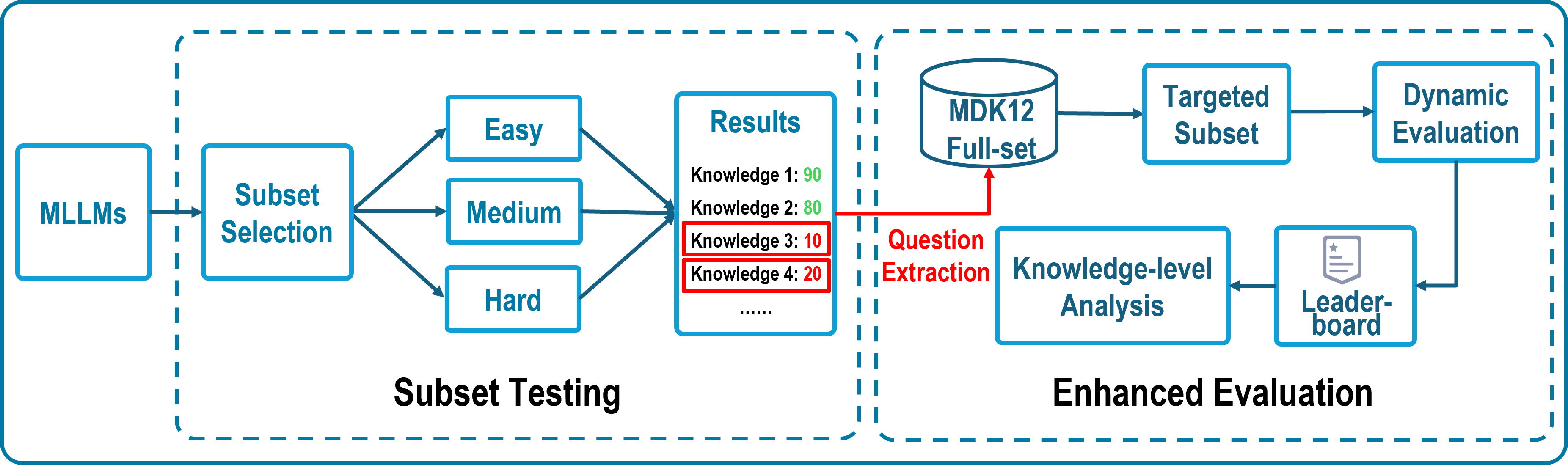}
    \caption{\textbf{Test logic of using subsets and the fullset data of MDK12-Bench progressively.}}
    \label{fig:testlogic}
\end{figure}

\subsection{Results on MDK12-mini}
\label{subsec:main_table_results}

We present the performance of baselines across easy, medium, and hard subsets in Table~\ref{tab:perf_tab1}. Gemini2-thinking achieves the highest overall accuracy of 59.4\%, notably excelling in Chemistry and Biology. Gemini2-flash follows closely with an overall accuracy of 57.2\%, surpassing Claude-3.7 and GPT-based series. QVQ-72B achieves the best performance (53.2\% overall accuracy) among open-source models, showing superior reasoning performance. InternVL2.5-MPO shows exceptional results in Biology at medium difficulty, though performance in other disciplines is mixed. InternVL2.5-8B consistently underperforms compared with other models. Overall, larger models generally perform better, while variability across different subjects is also observed, implying varied domain challenges. These results highlight the necessity for targeted improvements in discipline-specific multimodal reasoning.

Fig.~\ref{fig:knowledge_breakdown} presents the accuracy distribution with respect to specific knowledge points, aggregated across all subjects. Each question is annotated with one or more fine-grained knowledge labels. We highlight that models consistently achieve higher accuracy on frequently covered knowledge points in the training data, while systematically underperforming in areas such as advanced geometry and biochemical processes. The results highlight that our benchmark effectively helps identify specific knowledge gaps within models, enabling targeted in-depth evaluation of the full dataset and facilitating focused improvements in these weaker knowledge areas. Furthermore, Fig.~\ref{fig:year_breakdown} provides a year-by-year accuracy breakdown across different difficulty levels, highlighting temporal trends and potential shifts in model performance over time.
\textbf{Findings:} The results suggest that there is a possibility that earlier exam data is included more in the training set, therefore improving accuracy on these memorized similar questions.


\subsection{Results on Full MDK12-Bench}
\label{subsec:full_dataset}


We also evaluate each baseline on the entire dataset of 141.3K questions. 
Fig.~\ref{fig:full_set} summarizes the main results. Running on the full benchmark is computationally intensive; hence we prioritize the representative checkpoints from each model family. Results confirm that the trends seen in the subsets align well with the full set. \textbf{The Test Logic}: As illustrated in Fig.~\ref{fig:testlogic}, it is noted that our benchmark can serve as an additional testbed after detecting which knowledge the model failed. That is, the new proposed models are first tested on three subsets of MDK12-Bench, and provide knowledge-level performance. After gathering the key knowledge points that models do not perform well, the corresponding full-set data related to these key points can be extracted as a targeted subset for an enhanced evaluation. This is a preliminary dynamic evaluation process.

\begin{table*}[htbp]
\centering
\small
\setlength{\tabcolsep}{3.8pt}
\renewcommand{\arraystretch}{1}
\begin{tabular}{lcccccccccccc}
\hline
\multicolumn{1}{c}{\multirow{2}{*}{\textbf{Model}}} 
& \multicolumn{3}{c}{\textbf{Overall}} 
& \multicolumn{3}{c}{\textbf{Easy}} 
& \multicolumn{3}{c}{\textbf{Medium}} 
& \multicolumn{3}{c}{\textbf{Hard}} \\
\cmidrule(lr){2-4}\cmidrule(lr){5-7}\cmidrule(lr){8-10}\cmidrule(lr){11-13}
& Original & Dynamic & \textbf{\textcolor{red}{$\triangle$}}
& Original & Dynamic & \textbf{\textcolor{red}{$\triangle$}}
& Original & Dynamic & \textbf{\textcolor{red}{$\triangle$}}
& Original & Dynamic & \textbf{\textcolor{red}{$\triangle$}} \\
\hline
Gemini2-thinking 
& \textbf{58.1} & 41.6 & \textcolor{red}{\textbf{16.5}}
& \textbf{66.7} & 43.8 & \textcolor{red}{\textbf{22.9}}
& \textbf{57.0} & 44.8 & \textcolor{red}{\textbf{12.2}}
& \textbf{51.5} & 36.2 & \textcolor{red}{\textbf{15.3}} \\
Gemini2-flash 
& 56.4 & \textbf{47.0} & \textcolor{red}{9.4}
& 66.6 & \textbf{50.1} & \textcolor{red}{16.4}
& 54.7 & \textbf{46.1} & \textcolor{red}{8.6}
& 48.9 & \textbf{44.5} & \textcolor{red}{4.4} \\
Claude-3.7 
& 46.7 & 31.4 & \textcolor{red}{15.3}
& 49.2 & 32.3 & \textcolor{red}{16.9}
& 50.2 & 36.3 & \textcolor{red}{13.9}
& 40.5 & 25.2 & \textcolor{red}{15.3} \\
GPT-4o 
& 51.2 & 40.9 & \textcolor{red}{10.3}
& 54.1 & 35.7 & \textcolor{red}{18.5}
& 53.7 & 51.3 & \textcolor{red}{2.4}
& 35.4 & 34.8 & \textcolor{red}{0.6} \\
Qwen2-VL-7B-GRPO & 28.2 & 26.0 & \textcolor{red}{2.2} 
& 32.7 & 29.4 & \textcolor{red}{3.3} 
& 26.3 & 24.9 & \textcolor{red}{1.4} 
& 26.5 & 19.9 & \textcolor{red}{6.6}
\\
Qwen2-VL-7B 
& 27.3 & 26.1 & \textcolor{red}{1.2} 
& 31.8 & 34.6 & \textcolor{red}{-2.8} 
& 25.5 & 25.4 & \textcolor{red}{0.0} 
& 25.6 & 20.5 & \textcolor{red}{5.0} \\
InternVL2.5-8B 
& 41.7 & 26.1 & \textcolor{red}{15.6} 
& 48.5 & 23.5 & \textcolor{red}{25.0} 
& 44.1 & 27.5 & \textcolor{red}{16.6} 
& 38.4 & 30.8 & \textcolor{red}{7.7} \\
\hline
\end{tabular}
    \caption{\textbf{Accuracy on the original subset and the corresponding dynamic bootstrapped set.} The $\triangle$ denotes the accuracy fluctuation.}
    \label{tab:dynamic}
\end{table*}

\subsection{Dynamic Evaluation Results}
\label{subsec:dynamic_eval}

\begin{figure*}[t]
    \centering
    \includegraphics[trim=0.5 1.2cm 0 0, width=1.0\textwidth]{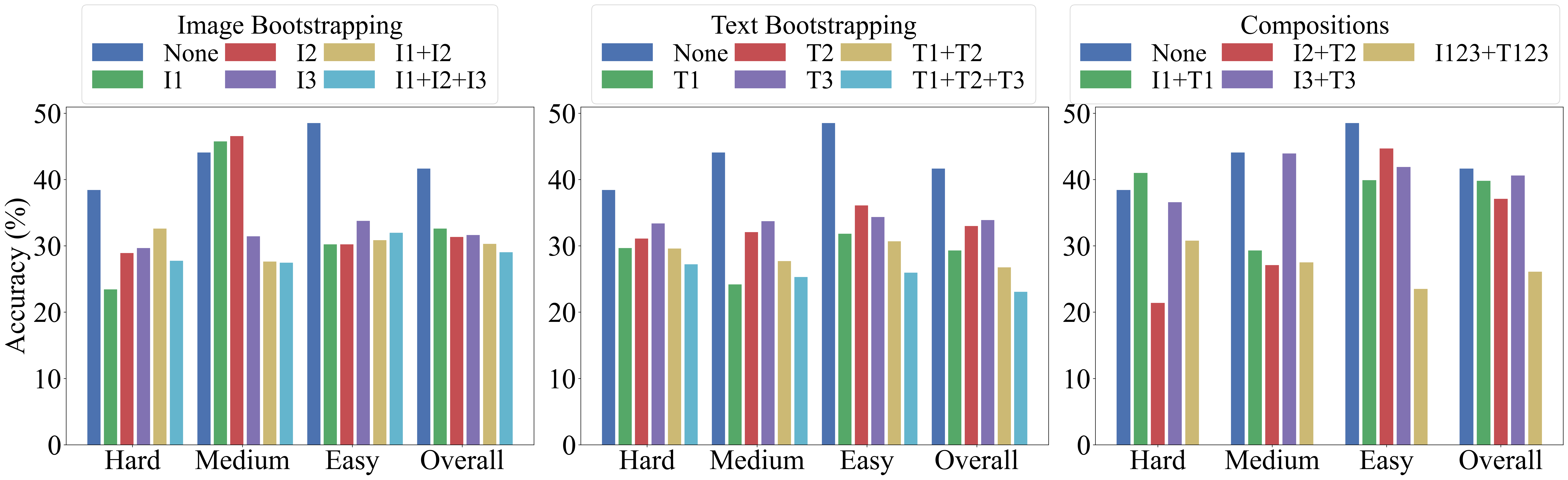}
    \caption{\textbf{Accuracy of InternVL2.5-8B on the sampled subset using different combinations of dynamic bootstrapping strategies.}}
    \label{fig:dynamic_eval}
\end{figure*}

\textbf{Main dynamic evaluation results.} We sampled 50\% multimodal instances from MDK12-mini as the original set (including 695 easy-level instances, 818 medium-level instances and 1124 hard-level instances. Then we applied all bootstrapping methods (3 textual and 3 visual) to create augmented test queries. Table~\ref{tab:dynamic} summarizes the results for each model under original vs.\ bootstrapped queries. The main insights are introduced as follows:

\textbf{1) Models show clear vulnerability to combined textual and visual bootstrapping:} The accuracy of MLLMs consistently dropped under combined bootstrapping, highlighting weaknesses in handling simultaneous context shifts. \textbf{2) Higher-performing models exhibit greater sensitivity to dynamic perturbations:} GPT-4o showed significant accuracy drops (10.3\%), emphasizing their heavy reliance on contextual reasoning rather than purely memorized knowledge. In contrast, Qwen2.5-VL-7B showed a relatively minor reduction (1.20\%), indicating reliance on certain question format distribution and, thus, greater stability under dynamic conditions. \textbf{3) High-order reasoning models are particularly sensitive to easier tasks under dynamic conditions:} Gemini2-think experienced a substantial accuracy decline (22.9\%) on easy tasks, suggesting these models rely significantly on precise contextual comprehension. It is easy to be disturbed by altered context outside the distribution.

\noindent\textbf{Ablation Study on Bootstrapping Combinations.} \label{subsec:ablation_study} 
We conduct an ablation study to analyze various random combinations of bootstrapping methods. Fig.~\ref{fig:dynamic_eval} reports the average accuracy fluctuation as we add different transformations. Results suggest the following findings:

\textbf{1) The composition bootstrapping strategy appears to have the strongest negative effect on model accuracy:} Significant reduction in accuracy is observed when multiple image (I) or textual (T) perturbations are combined. Particularly, the combination of I1+I2+I3 and T1+T2+T3 contributed to the strongest overall reductions, including 12.7\% and 18.6\% respective reductions in image and text.
\textbf{2) Transformations in text produce a stronger reduction in accuracy than images:} A single textual perturbation (e.g., T1/T2/T3) consistently produces lower accuracy than a single visual perturbation (I1/I2/I3) indicating models' greater reliance on the textual context during reasoning.
\textbf{3) The hard tasks are less affected by the combined perturbations compared to easier tasks.} For example, the accuracy on hard tasks typically remains stable or even slightly increases with certain compositions (e.g., I1+T1). It is possible that models inherently reason harder to solve hard tasks than simple memorization. These observations indicate the critical need for developing models that are capable of maintaining robust multimodal comprehension and reasoning under dynamic and diverse conditions.







\section{Conclusion}

We introduce MDK12-Bench, a comprehensive multimodal benchmark designed to evaluate the reasoning abilities of MLLMs across diverse real-world K-12 tasks. By covering 141K questions that span multiple disciplines and grade levels, structuring the knowledge system through fine-grained knowledge-point annotations and detailed answer explanations, MDK12-Bench fills critical gaps present in existing benchmarks, such as limited data scale, narrow domain coverage, and unstructured knowledge representation. We further proposed a dynamic evaluation framework employing multiple textual and visual bootstrapping strategies to mitigate data contamination, ensuring robust, reliable, and fair assessment. Experimental results revealed significant limitations in current state-of-the-art MLLMs, particularly highlighting their sensitivity to contextual changes and task complexity. Our insights highlight the necessity for enhanced multimodal contextual comprehension and reasoning abilities. 
Future work may focus on improving models' resilience to dynamic perturbations of question formats, thereby paving the way toward more robust multimodal reasoning models.

{
    \small
    \bibliographystyle{ieeenat_fullname}
    \bibliography{main}

\begin{thebibliography}{56}
\providecommand{\natexlab}[1]{#1}
\providecommand{\url}[1]{\texttt{#1}}
\expandafter\ifx\csname urlstyle\endcsname\relax
  \providecommand{\doi}[1]{doi: #1}\else
  \providecommand{\doi}{doi: \begingroup \urlstyle{rm}\Url}\fi

\bibitem[Achiam et~al.(2023)Achiam, Adler, Agarwal, Ahmad, Akkaya, Aleman, Almeida, Altenschmidt, Altman, Anadkat, et~al.]{openai2023gpt}
Josh Achiam, Steven Adler, Sandhini Agarwal, Lama Ahmad, Ilge Akkaya, Florencia~Leoni Aleman, Diogo Almeida, Janko Altenschmidt, Sam Altman, Shyamal Anadkat, et~al.
\newblock {GPT-4} technical report.
\newblock \emph{arXiv preprint arXiv:2303.08774}, 2023.

\bibitem[Alayrac et~al.(2022)Alayrac, Donahue, Luc, Miech, Barr, Hasson, Lenc, Mensch, Millican, Reynolds, et~al.]{alayrac2022flamingo}
Jean-Baptiste Alayrac, Jeff Donahue, Pauline Luc, Antoine Miech, Iain Barr, Yana Hasson, Karel Lenc, Arthur Mensch, Katherine Millican, Malcolm Reynolds, et~al.
\newblock Flamingo: a visual language model for few-shot learning.
\newblock \emph{Advances in neural information processing systems}, 35:\penalty0 23716--23736, 2022.

\bibitem[Anthropic(2023)]{claud3}
Anthropic.
\newblock The claude 3 model family: Opus, sonnet, haiku.
\newblock 2023.

\bibitem[Bai et~al.(2025)Bai, Chen, Liu, Wang, Ge, Song, Dang, Wang, Wang, Tang, et~al.]{bai2025qwen2}
Shuai Bai, Keqin Chen, Xuejing Liu, Jialin Wang, Wenbin Ge, Sibo Song, Kai Dang, Peng Wang, Shijie Wang, Jun Tang, et~al.
\newblock Qwen2. 5-vl technical report.
\newblock \emph{arXiv preprint arXiv:2502.13923}, 2025.

\bibitem[Baral et~al.(2024)Baral, Lucy, Knight, Ng, Soldaini, Heffernan, and Lo]{baral24drawedumath}
Sami Baral, Li Lucy, Ryan Knight, Alice Ng, Luca Soldaini, Neil Heffernan, and Kyle Lo.
\newblock Drawedumath: Evaluating vision language models with expert-annotated students’ hand-drawn math images.
\newblock In \emph{The 4th Workshop on Mathematical Reasoning and AI at NeurIPS'24}, 2024.

\bibitem[{Black Forest Labs}(2024)]{blackforestlabs_flux}
{Black Forest Labs}.
\newblock Flux.
\newblock \url{https://github.com/black-forest-labs/flux}, 2024.
\newblock Accessed: 2024-11-05.

\bibitem[Chen et~al.(2024{\natexlab{a}})Chen, Wang, Xu, Cao, Fang, and Lin]{chen2024systematic}
Eason Chen, Danyang Wang, Luyi Xu, Chen Cao, Xiao Fang, and Jionghao Lin.
\newblock A systematic review on prompt engineering in large language models for k-12 stem education.
\newblock \emph{arXiv preprint arXiv:2410.11123}, 2024{\natexlab{a}}.

\bibitem[Chen et~al.(2024{\natexlab{b}})Chen, Wang, Cao, Liu, Gao, Cui, Zhu, Ye, Tian, Liu, et~al.]{chen2024expandinginternvl2.5}
Zhe Chen, Weiyun Wang, Yue Cao, Yangzhou Liu, Zhangwei Gao, Erfei Cui, Jinguo Zhu, Shenglong Ye, Hao Tian, Zhaoyang Liu, et~al.
\newblock Expanding performance boundaries of open-source multimodal models with model, data, and test-time scaling.
\newblock \emph{arXiv preprint arXiv:2412.05271}, 2024{\natexlab{b}}.

\bibitem[Cobbe et~al.(2021)Cobbe, Kosaraju, Bavarian, Chen, Jun, Kaiser, Plappert, Tworek, Hilton, Nakano, et~al.]{cobbe2021training}
Karl Cobbe, Vineet Kosaraju, Mohammad Bavarian, Mark Chen, Heewoo Jun, Lukasz Kaiser, Matthias Plappert, Jerry Tworek, Jacob Hilton, Reiichiro Nakano, et~al.
\newblock Training verifiers to solve math word problems.
\newblock \emph{arXiv preprint arXiv:2110.14168}, 2021.

\bibitem[Das et~al.(2024)Das, Hristov, Li, Dimitrov, Koychev, and Nakov]{das2024exams}
Rocktim~Jyoti Das, Simeon~Emilov Hristov, Haonan Li, Dimitar~Iliyanov Dimitrov, Ivan Koychev, and Preslav Nakov.
\newblock Exams-v: A multi-discipline multilingual multimodal exam benchmark for evaluating vision language models.
\newblock \emph{arXiv preprint arXiv:2403.10378}, 2024.

\bibitem[EvolvingLMMs-Lab()]{openr1multimodal}
EvolvingLMMs-Lab.
\newblock open-r1-multimodal: A fork to add multimodal model training to open-r1.
\newblock \url{https://github.com/EvolvingLMMs-Lab/open-r1-multimodal}.
\newblock Accessed: 2025-03-08.

\bibitem[Guo et~al.(2025)Guo, Yang, Zhang, Song, Zhang, Xu, Zhu, Ma, Wang, Bi, et~al.]{guo2025deepseek}
Daya Guo, Dejian Yang, Haowei Zhang, Junxiao Song, Ruoyu Zhang, Runxin Xu, Qihao Zhu, Shirong Ma, Peiyi Wang, Xiao Bi, et~al.
\newblock Deepseek-r1: Incentivizing reasoning capability in llms via reinforcement learning.
\newblock \emph{arXiv preprint arXiv:2501.12948}, 2025.

\bibitem[Hao et~al.(2025)Hao, Gu, Wang, Li, Yang, Wang, and Cheng]{hao2025can}
Yunzhuo Hao, Jiawei Gu, Huichen~Will Wang, Linjie Li, Zhengyuan Yang, Lijuan Wang, and Yu Cheng.
\newblock Can mllms reason in multimodality? emma: An enhanced multimodal reasoning benchmark.
\newblock \emph{arXiv preprint arXiv:2501.05444}, 2025.

\bibitem[Huang and Zhang(2024)]{huang2024survey}
Jiaxing Huang and Jingyi Zhang.
\newblock A survey on evaluation of multimodal large language models.
\newblock \emph{arXiv preprint arXiv:2408.15769}, 2024.

\bibitem[Hudson and Manning(2019)]{hudson2019gqa}
Drew~A Hudson and Christopher~D Manning.
\newblock Gqa: A new dataset for real-world visual reasoning and compositional question answering.
\newblock In \emph{Proceedings of the IEEE/CVF conference on computer vision and pattern recognition}, pages 6700--6709, 2019.

\bibitem[Jiang et~al.(2024)Jiang, Zheng, Zhang, Jin, Yuan, and Liu]{jiang2024med}
Songtao Jiang, Tuo Zheng, Yan Zhang, Yeying Jin, Li Yuan, and Zuozhu Liu.
\newblock Med-moe: Mixture of domain-specific experts for lightweight medical vision-language models.
\newblock In \emph{Findings of the Association for Computational Linguistics: EMNLP 2024}, pages 3843--3860, 2024.

\bibitem[Kil et~al.()Kil, Mai, Lee, Chowdhury, Wang, Cheng, Wang, Liu, and Chao]{kilmllm}
Jihyung Kil, Zheda Mai, Justin Lee, Arpita Chowdhury, Zihe Wang, Kerrie Cheng, Lemeng Wang, Ye Liu, and Wei-Lun Chao.
\newblock Mllm-compbench: A comparative reasoning benchmark for multimodal llms.
\newblock In \emph{The Thirty-eight Conference on Neural Information Processing Systems Datasets and Benchmarks Track}.

\bibitem[Li et~al.(2022)Li, Li, Xiong, and Hoi]{li2022blip}
Junnan Li, Dongxu Li, Caiming Xiong, and Steven Hoi.
\newblock Blip: Bootstrapping language-image pre-training for unified vision-language understanding and generation.
\newblock In \emph{International conference on machine learning}, pages 12888--12900. PMLR, 2022.

\bibitem[Li et~al.(2024)Li, Chen, Shi, Xiao, and Chen]{li2024survey}
Lin Li, Guikun Chen, Hanrong Shi, Jun Xiao, and Long Chen.
\newblock A survey on multimodal benchmarks: In the era of large ai models.
\newblock \emph{arXiv preprint arXiv:2409.18142}, 2024.

\bibitem[Liu et~al.(2024{\natexlab{a}})Liu, Li, Wu, and Lee]{liu2023llava}
Haotian Liu, Chunyuan Li, Qingyang Wu, and Yong~Jae Lee.
\newblock Visual instruction tuning.
\newblock \emph{Advances in neural information processing systems}, 36, 2024{\natexlab{a}}.

\bibitem[Liu et~al.(2024{\natexlab{b}})Liu, Liu, Wang, Xu, Chen, and Cheng]{liu2024k}
Jun Liu, Zile Liu, Cong Wang, Yanhua Xu, Jiayu Chen, and Yichun Cheng.
\newblock K-12 students' higher-order thinking skills: Conceptualization, components, and evaluation indicators.
\newblock \emph{Thinking Skills and Creativity}, 52:\penalty0 101551, 2024{\natexlab{b}}.

\bibitem[Liu et~al.(2024{\natexlab{c}})Liu, Duan, Zhang, Li, Zhang, Zhao, Yuan, Wang, He, Liu, et~al.]{liu2024mmbench}
Yuan Liu, Haodong Duan, Yuanhan Zhang, Bo Li, Songyang Zhang, Wangbo Zhao, Yike Yuan, Jiaqi Wang, Conghui He, Ziwei Liu, et~al.
\newblock Mmbench: Is your multi-modal model an all-around player?
\newblock In \emph{European conference on computer vision}, pages 216--233. Springer, 2024{\natexlab{c}}.

\bibitem[Lohman and Lakin(2011)]{lohman2011intelligence}
David~F Lohman and Joni~M Lakin.
\newblock Intelligence and reasoning.
\newblock \emph{The Cambridge handbook of intelligence}, pages 419--441, 2011.

\bibitem[Lu et~al.(2023)Lu, Bansal, Xia, Liu, Li, Hajishirzi, Cheng, Chang, Galley, and Gao]{lumathvista}
Pan Lu, Hritik Bansal, Tony Xia, Jiacheng Liu, Chunyuan Li, Hannaneh Hajishirzi, Hao Cheng, Kai-Wei Chang, Michel Galley, and Jianfeng Gao.
\newblock Mathvista: Evaluating mathematical reasoning of foundation models in visual contexts.
\newblock In \emph{The Twelfth International Conference on Learning Representations}, 2023.

\bibitem[Meng et~al.()Meng, Li, Wang, Lu, Tian, Yang, Liao, Zhu, Dai, Qiao, et~al.]{mengmmiu}
Fanqing Meng, Chuanhao Li, Jin Wang, Quanfeng Lu, Hao Tian, Tianshuo Yang, Jiaqi Liao, Xizhou Zhu, Jifeng Dai, Yu Qiao, et~al.
\newblock Mmiu: Multimodal multi-image understanding for evaluating large vision-language models.
\newblock In \emph{The Thirteenth International Conference on Learning Representations}.

\bibitem[Moor et~al.(2023)Moor, Huang, Wu, Yasunaga, Dalmia, Leskovec, Zakka, Reis, and Rajpurkar]{moor2023med}
Michael Moor, Qian Huang, Shirley Wu, Michihiro Yasunaga, Yash Dalmia, Jure Leskovec, Cyril Zakka, Eduardo~Pontes Reis, and Pranav Rajpurkar.
\newblock Med-flamingo: a multimodal medical few-shot learner.
\newblock In \emph{Machine Learning for Health (ML4H)}, pages 353--367. PMLR, 2023.

\bibitem[Morris et~al.(2023)Morris, Sohl-Dickstein, Fiedel, Warkentin, Dafoe, Faust, Farabet, and Legg]{morris2023levels}
Meredith~Ringel Morris, Jascha Sohl-Dickstein, Noah Fiedel, Tris Warkentin, Allan Dafoe, Aleksandra Faust, Clement Farabet, and Shane Legg.
\newblock Levels of agi: Operationalizing progress on the path to agi.
\newblock \emph{arXiv preprint arXiv:2311.02462}, 2023.

\bibitem[OpenAI(2023)]{gpt4v}
OpenAI.
\newblock Gpt-4v(ision) system card.
\newblock 2023.

\bibitem[OpenAI(2024{\natexlab{a}})]{openai2024gpt4o}
OpenAI.
\newblock Gpt-4o: A multimodal language model, 2024{\natexlab{a}}.
\newblock Accessed: 2025-03-08.

\bibitem[OpenAI(2024{\natexlab{b}})]{openai2024gpto1mini}
OpenAI.
\newblock Gpt-o1-mini: A multimodal language model, 2024{\natexlab{b}}.
\newblock Accessed: 2025-03-08.

\bibitem[OpenAI(2025)]{openai2025gpto3mini}
OpenAI.
\newblock Gpt-o3-mini: A cost-effective reasoning model, 2025.
\newblock Accessed: 2025-03-08.

\bibitem[Peng et~al.(2024)Peng, Fu, Gao, Zhong, Fu, and Tang]{peng2024multimath}
Shuai Peng, Di Fu, Liangcai Gao, Xiuqin Zhong, Hongguang Fu, and Zhi Tang.
\newblock Multimath: Bridging visual and mathematical reasoning for large language models.
\newblock \emph{arXiv preprint arXiv:2409.00147}, 2024.

\bibitem[Radford et~al.(2021)Radford, Kim, Hallacy, Ramesh, Goh, Agarwal, Sastry, Askell, Mishkin, Clark, et~al.]{radford2021learning}
Alec Radford, Jong~Wook Kim, Chris Hallacy, Aditya Ramesh, Gabriel Goh, Sandhini Agarwal, Girish Sastry, Amanda Askell, Pamela Mishkin, Jack Clark, et~al.
\newblock Learning transferable visual models from natural language supervision.
\newblock In \emph{International conference on machine learning}, pages 8748--8763. PmLR, 2021.

\bibitem[Rajabi and Kosecka()]{rajabi2024gsr}
Navid Rajabi and Jana Kosecka.
\newblock Gsr-bench: A benchmark for grounded spatial reasoning evaluation via multimodal llms.
\newblock In \emph{NeurIPS 2024 Workshop on Compositional Learning: Perspectives, Methods, and Paths Forward}.

\bibitem[Saikh et~al.(2022)Saikh, Ghosal, Mittal, Ekbal, and Bhattacharyya]{saikh2022scienceqa}
Tanik Saikh, Tirthankar Ghosal, Amish Mittal, Asif Ekbal, and Pushpak Bhattacharyya.
\newblock Scienceqa: A novel resource for question answering on scholarly articles.
\newblock \emph{International Journal on Digital Libraries}, 23\penalty0 (3):\penalty0 289--301, 2022.

\bibitem[Shi et~al.(2024)Shi, Hu, Bin, Liu, Yang, Ng, Bing, and Lee]{shi2024math}
Wenhao Shi, Zhiqiang Hu, Yi Bin, Junhua Liu, Yang Yang, See~Kiong Ng, Lidong Bing, and Roy Lee.
\newblock Math-llava: Bootstrapping mathematical reasoning for multimodal large language models.
\newblock In \emph{Findings of the Association for Computational Linguistics: EMNLP 2024}, pages 4663--4680, 2024.

\bibitem[Sternberg(1982)]{sternberg1982reasoning}
Robert~J Sternberg.
\newblock Reasoning, problem solving, and intelligence.
\newblock \emph{Handbook of human intelligence}, pages 225--307, 1982.

\bibitem[Sun et~al.(2024)Sun, Wu, Zhu, Zheng, Chen, Zhang, Zhang, Wan, Lan, Zheng, et~al.]{sun2024pathmmu}
Yuxuan Sun, Hao Wu, Chenglu Zhu, Sunyi Zheng, Qizi Chen, Kai Zhang, Yunlong Zhang, Dan Wan, Xiaoxiao Lan, Mengyue Zheng, et~al.
\newblock Pathmmu: A massive multimodal expert-level benchmark for understanding and reasoning in pathology.
\newblock In \emph{European Conference on Computer Vision}, pages 56--73. Springer, 2024.

\bibitem[Tanaka et~al.(2023)Tanaka, Nishida, Nishida, Hasegawa, Saito, and Saito]{tanaka2023slidevqa}
Ryota Tanaka, Kyosuke Nishida, Kosuke Nishida, Taku Hasegawa, Itsumi Saito, and Kuniko Saito.
\newblock Slidevqa: A dataset for document visual question answering on multiple images.
\newblock In \emph{Proceedings of the AAAI Conference on Artificial Intelligence}, pages 13636--13645, 2023.

\bibitem[Team et~al.(2023)Team, Anil, Borgeaud, Alayrac, Yu, Soricut, Schalkwyk, Dai, Hauth, Millican, et~al.]{team2023gemini}
Gemini Team, Rohan Anil, Sebastian Borgeaud, Jean-Baptiste Alayrac, Jiahui Yu, Radu Soricut, Johan Schalkwyk, Andrew~M Dai, Anja Hauth, Katie Millican, et~al.
\newblock Gemini: a family of highly capable multimodal models.
\newblock \emph{arXiv preprint arXiv:2312.11805}, 2023.

\bibitem[Team(2024)]{qvq-72b-preview}
Qwen Team.
\newblock Qvq: To see the world with wisdom, 2024.

\bibitem[Touvron et~al.(2023)Touvron, Lavril, Izacard, Martinet, Lachaux, Lacroix, Rozi{\`e}re, Goyal, Hambro, Azhar, et~al.]{touvron2023llama}
Hugo Touvron, Thibaut Lavril, Gautier Izacard, Xavier Martinet, Marie-Anne Lachaux, Timoth{\'e}e Lacroix, Baptiste Rozi{\`e}re, Naman Goyal, Eric Hambro, Faisal Azhar, et~al.
\newblock {LLaMA}: Open and efficient foundation language models.
\newblock \emph{arXiv preprint arXiv:2302.13971}, 2023.

\bibitem[Wang et~al.(2025)Wang, Pan, Shi, Lu, Ren, Zhou, Zhan, and Li]{wang2025measuring}
Ke Wang, Junting Pan, Weikang Shi, Zimu Lu, Houxing Ren, Aojun Zhou, Mingjie Zhan, and Hongsheng Li.
\newblock Measuring multimodal mathematical reasoning with math-vision dataset.
\newblock \emph{Advances in Neural Information Processing Systems}, 37:\penalty0 95095--95169, 2025.

\bibitem[Wang et~al.(2024)Wang, Chen, Wang, Cao, Liu, Gao, Zhu, Zhu, Lu, Qiao, and Dai]{wang2024mpo}
Weiyun Wang, Zhe Chen, Wenhai Wang, Yue Cao, Yangzhou Liu, Zhangwei Gao, Jinguo Zhu, Xizhou Zhu, Lewei Lu, Yu Qiao, and Jifeng Dai.
\newblock Enhancing the reasoning ability of multimodal large language models via mixed preference optimization.
\newblock \emph{arXiv preprint arXiv:2411.10442}, 2024.

\bibitem[Wei et~al.(2022)Wei, Wang, Schuurmans, Bosma, Xia, Chi, Le, Zhou, et~al.]{wei2022chain}
Jason Wei, Xuezhi Wang, Dale Schuurmans, Maarten Bosma, Fei Xia, Ed Chi, Quoc~V Le, Denny Zhou, et~al.
\newblock Chain-of-thought prompting elicits reasoning in large language models.
\newblock \emph{Advances in neural information processing systems}, 35:\penalty0 24824--24837, 2022.

\bibitem[Xia et~al.(2025)Xia, Chen, Tian, Yangrui, Hou, Xu, Wu, Fan, Zhou, Zhu, et~al.]{xiacares}
Peng Xia, Ze Chen, Juanxi Tian, Gong Yangrui, Ruibo Hou, Yue Xu, Zhenbang Wu, Zhiyuan Fan, Yiyang Zhou, Kangyu Zhu, et~al.
\newblock Cares: A comprehensive benchmark of trustworthiness in medical vision language models.
\newblock In \emph{The Thirty-eight Conference on Neural Information Processing Systems Datasets and Benchmarks Track}, 2025.

\bibitem[Yang et~al.(2024{\natexlab{a}})Yang, Ge, Li, Chen, Ge, Shan, and Chen]{yang2024seed}
Shuai Yang, Yuying Ge, Yang Li, Yukang Chen, Yixiao Ge, Ying Shan, and Yingcong Chen.
\newblock Seed-story: Multimodal long story generation with large language model.
\newblock \emph{arXiv preprint arXiv:2407.08683}, 2024{\natexlab{a}}.

\bibitem[Yang et~al.(2024{\natexlab{b}})Yang, Zhang, Shao, Zhang, Bin, Wang, and Luo]{yang2024dynamic}
Yue Yang, Shuibai Zhang, Wenqi Shao, Kaipeng Zhang, Yi Bin, Yu Wang, and Ping Luo.
\newblock Dynamic multimodal evaluation with flexible complexity by vision-language bootstrapping.
\newblock \emph{arXiv preprint arXiv:2410.08695}, 2024{\natexlab{b}}.

\bibitem[Ying et~al.()Ying, Meng, Wang, Li, Lin, Yang, Zhang, Zhang, Lin, Liu, et~al.]{yingmmt}
Kaining Ying, Fanqing Meng, Jin Wang, Zhiqian Li, Han Lin, Yue Yang, Hao Zhang, Wenbo Zhang, Yuqi Lin, Shuo Liu, et~al.
\newblock Mmt-bench: A comprehensive multimodal benchmark for evaluating large vision-language models towards multitask agi.
\newblock In \emph{Forty-first International Conference on Machine Learning}.

\bibitem[Yue et~al.(2024)Yue, Ni, Zhang, Zheng, Liu, Zhang, Stevens, Jiang, Ren, Sun, et~al.]{yue2024mmmu}
Xiang Yue, Yuansheng Ni, Kai Zhang, Tianyu Zheng, Ruoqi Liu, Ge Zhang, Samuel Stevens, Dongfu Jiang, Weiming Ren, Yuxuan Sun, et~al.
\newblock {MMMU}: A massive multi-discipline multimodal understanding and reasoning benchmark for expert {AGI}.
\newblock In \emph{Proceedings of the IEEE/CVF Conference on Computer Vision and Pattern Recognition}, pages 9556--9567, 2024.

\bibitem[Zelikman et~al.(2024{\natexlab{a}})Zelikman, Harik, Shao, Jayasiri, Haber, and Goodman]{zelikman2024quiet}
Eric Zelikman, Georges Harik, Yijia Shao, Varuna Jayasiri, Nick Haber, and Noah~D Goodman.
\newblock {Quiet-STaR}: Language models can teach themselves to think before speaking.
\newblock \emph{arXiv preprint arXiv:2403.09629}, 2024{\natexlab{a}}.

\bibitem[Zelikman et~al.(2024{\natexlab{b}})Zelikman, Wu, Mu, and Goodman]{zelikman2024star}
Eric Zelikman, YH Wu, Jesse Mu, and Noah~D Goodman.
\newblock Star: Self-taught reasoner bootstrapping reasoning with reasoning.
\newblock In \emph{Proc. the 36th International Conference on Neural Information Processing Systems}, 2024{\natexlab{b}}.

\bibitem[Zhang et~al.(2024)Zhang, Jiang, Zhang, Lin, Guo, Qiu, Zhou, Lu, Chang, Qiao, et~al.]{zhang2024mathverse}
Renrui Zhang, Dongzhi Jiang, Yichi Zhang, Haokun Lin, Ziyu Guo, Pengshuo Qiu, Aojun Zhou, Pan Lu, Kai-Wei Chang, Yu Qiao, et~al.
\newblock Mathverse: Does your multi-modal llm truly see the diagrams in visual math problems?
\newblock In \emph{European Conference on Computer Vision}, pages 169--186. Springer, 2024.

\bibitem[Zhong et~al.(2024)Zhong, Feng, Xiong, Cheng, Zhao, He, Bian, and Wang]{zhong2024dpo}
Han Zhong, Guhao Feng, Wei Xiong, Xinle Cheng, Li Zhao, Di He, Jiang Bian, and Liwei Wang.
\newblock Dpo meets ppo: Reinforced token optimization for rlhf.
\newblock \emph{arXiv preprint arXiv:2404.18922}, 2024.

\bibitem[Zhou et~al.(2024)Zhou, Peng, Song, Li, Xu, Yang, Guo, Zhang, Lin, He, et~al.]{zhou2024gate}
Pengfei Zhou, Xiaopeng Peng, Jiajun Song, Chuanhao Li, Zhaopan Xu, Yue Yang, Ziyao Guo, Hao Zhang, Yuqi Lin, Yefei He, et~al.
\newblock Gate opening: A comprehensive benchmark for judging open-ended interleaved image-text generation.
\newblock \emph{arXiv preprint arXiv:2411.18499}, 2024.

\bibitem[Zhu et~al.(2024)Zhu, Wang, Zhao, Xu, and Xie]{zhu2024dynamic}
Kaijie Zhu, Jindong Wang, Qinlin Zhao, Ruochen Xu, and Xing Xie.
\newblock Dynamic evaluation of large language models by meta probing agents.
\newblock In \emph{Forty-first International Conference on Machine Learning}, 2024.

\end{thebibliography}
}


\end{document}


\maketitle

\noindent

In this supplementary material, we provide additional information, discussions, and results in support of the primary text, which are organized as follows:

\begin{table*}[!htbp]
  \centering
  \small
  \begin{tabularx}{0.99\textwidth}{@{}
    >{\raggedright\arraybackslash}p{3.5cm} 
    *{10}{Y}                         
    @{ \hspace{0.08cm} }
  @{}}
    \toprule
    \multicolumn{1}{l}{\multirow{2}{*}{\hspace{-1.5mm}\textbf{Models}}} 
      & \multicolumn{10}{c}{\textbf{Years}} \\
    \cmidrule(lr){2-11}
& 2016 & 2017 & 2018 & 2019 & 2020 & 2021 & 2022 & 2023 & 2024 & 2025 \\
\midrule
\textbf{Easy} & & & & & & & & & & \\
QVQ-72B & 64.8 & 61.0 & 60.1 & 55.1 & 65.4 & 51.8 & 52.1 & 56.0 & 58.0 & 42.0 \\
Qwen2.5-VL-72B & 70.0 & 66.4 & 57.6 & 53.2 & 69.4 & 64.5 & 55.7 & 54.8 & 52.8 & 44.5 \\
GPT-4o & 70.0 & 61.9 & 54.5 & 53.6 & 73.0 & 69.7 & 52.7 & 55.5 & 52.6 & 42.4 \\
Gemini2-thinking & 48.4 & 78.4 & 76.0 & 65.9 & 72.5 & 64.5 & 66.3 & 66.9 & 69.2 & 60.8 \\
Gemini-2-flash & 63.3 & 76.9 & 71.7 & 64.3 & 75.0 & 56.8 & 62.4 & 61.9 & 71.8 & 65.7 \\
\midrule
\textbf{Medium} & & & & & & & & & & \\
QVQ-72B & 61.8 & 62.1 & 62.8 & 59.1 & 51.7 & 51.7 & 50.3 & 52.1 & 47.7 & 32.3 \\
Qwen2.5-VL-72B & 62.6 & 59.2 & 70.8 & 50.5 & 45.0 & 37.1 & 52.1 & 45.1 & 46.8 & 70.5 \\
GPT-4o & 76.2 & 64.6 & 60.1 & 46.8 & 70.3 & 61.9 & 60.0 & 57.1 & 65.8 & 49.7 \\
Gemini2-thinking & 66.0 & 64.5 & 72.4 & 67.0 & 60.3 & 53.3 & 59.3 & 57.4 & 53.6 & 54.8 \\
Gemini-2-flash & 68.5 & 67.2 & 69.8 & 63.4 & 65.7 & 58.9 & 61.2 & 59.8 & 58.3 & 57.6 \\
\midrule
\textbf{Hard} & & & & & & & & & & \\
QVQ-72B & 56.7 & 70.7 & 52.8 & 47.2 & 44.9 & 44.1 & 50.9 & 46.0 & 44.7 & 55.6 \\
Qwen2.5-VL-72B & 36.3 & 53.3 & 47.2 & 39.5 & 46.2 & 36.8 & 45.0 & 41.6 & 37.3 & 40.6 \\
GPT-4o & 58.5 & 70.4 & 52.7 & 45.7 & 42.8 & 33.1 & 45.6 & 38.7 & 32.8 & 48.7 \\
Gemini2-thinking & 63.1 & 66.3 & 56.6 & 54.6 & 51.8 & 51.7 & 50.8 & 46.2 & 48.3 & 60.8 \\
Gemini-2-flash & 59.8 & 68.5 & 54.3 & 50.2 & 48.9 & 47.5 & 48.6 & 44.3 & 45.7 & 56.4 \\
\bottomrule
\end{tabularx}
\vspace{-0.2cm}
\caption{\textbf{Model performance across different difficulty levels (easy, medium, hard) over the years 2016 to 2025.} Each cell represents the accuracy of the corresponding model in a specific year.}
\label{tab:perf_years}
\vspace{-0.1cm}
\end{table*}

\begin{algorithm}[ht]
\caption{Dynamic Bootstrapping \& Evaluation}
\label{alg:example}

\begin{algorithmic}[1]
\Require 
  \(\mathcal{S} = \{(Q,I,A)\}\)\Comment{Original dataset: each sample has question \(Q\), image \(I\), and ground-truth \(A\)}
  \Statex \(\mathcal{T}_{\mathrm{text}}\)\Comment{Set of textual transformations (e.g., word substitution, paraphrasing and type permuting)}
  \Statex \(\mathcal{T}_{\mathrm{vis}}\)\Comment{Set of visual transformations (e.g., expansion, color shift and style transfer)}
  \Statex \(M\)\Comment{The MLLM to be evaluated}
\Ensure 
  \(\text{Scores}\)\Comment{Final evaluated scores}

\State \(\mathcal{S}_{aug} \gets \varnothing\) \Comment{Augmented dataset (initially empty)}

\For{each \((Q,I,A) \in \mathcal{S}\)}
  \State \(\mathcal{Q'} \gets \varnothing,\; \mathcal{I'} \gets \varnothing\) \Comment{Sets for valid textual/visual variants}
  \For{each \(t \in \mathcal{T}_{\mathrm{text}}\)}
    \State \(Q_t \gets \mathrm{TextTransform}(Q, t)\) \Comment{Transform question \(Q\) with method \(t\)}
    \If{\(\mathrm{Judge}(Q_t, A) = \mathrm{True}\)}
      \State \(\mathcal{Q'} \gets \mathcal{Q'} \cup \{Q_t\}\)
    \EndIf
  \EndFor

  \For{each \(v \in \mathcal{T}_{\mathrm{vis}}\)}
    \State \(I_v \gets \mathrm{VisTransform}(I, v)\) \Comment{Transform image \(I\) with method \(v\)}
    \If{\(\mathrm{Judge}(I_v, A) = \mathrm{True}\)}
      \State \(\mathcal{I'} \gets \mathcal{I'} \cup \{I_v\}\)
    \EndIf
  \EndFor

  \State \(\mathcal{Q'} \gets \mathcal{Q'} \cup \{Q\}\) \Comment{Include original question}
  \State \(\mathcal{I'} \gets \mathcal{I'} \cup \{I\}\) \Comment{Include original image}

  \For{each \(Q^* \in \mathcal{Q'}\)}
    \For{each \(I^* \in \mathcal{I'}\)}
      \State \(\mathcal{S}_{aug} \gets \mathcal{S}_{aug} \cup \{(Q^*, I^*, A)\}\) \Comment{Construct new sample}
    \EndFor
  \EndFor
\EndFor

\State \(\text{scores} \gets \varnothing\) \Comment{List of model scores}
\For{each \((Q_d, I_d, A) \in \mathcal{S}_{aug}\)}
  \State \(r \gets M(Q_d, I_d)\) \Comment{Model output (raw response)}
  \State \(a \gets \mathrm{Parse}(r)\) \Comment{Extract final predicted answer from \(r\)}
  \If{\(a = A\)}
    \State \(s \gets 1\) \Comment{Exact match}
  \Else
    \State \(s \gets \mathrm{Partial}(a,A)\) \Comment{Partial-credit scoring}
  \EndIf
  \State \(\text{scores} \gets \text{scores} \cup \{s\}\)
\EndFor

\State \Return \(\text{scores}\)

\end{algorithmic}
\end{algorithm}

\begin{figure*}
    \centering
    \centering
    \includegraphics[trim=0cm 0.5cm 0 0, width=0.9\textwidth]{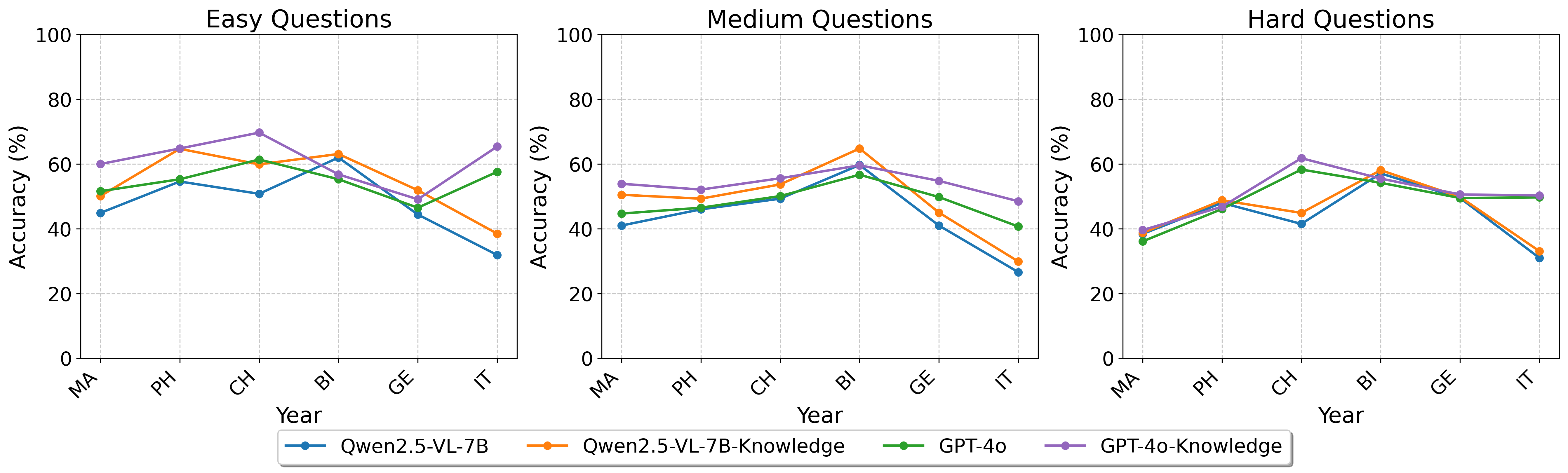}
    \caption{Experiments on Knowledge-Point Parsing Enhancement of two models at different levels of difficulty.}
    \label{fig:kg}
\end{figure*}

\begin{figure*}
    \centering
    \centering
    \includegraphics[trim=0cm 0.5cm 0 0, width=0.9\textwidth]{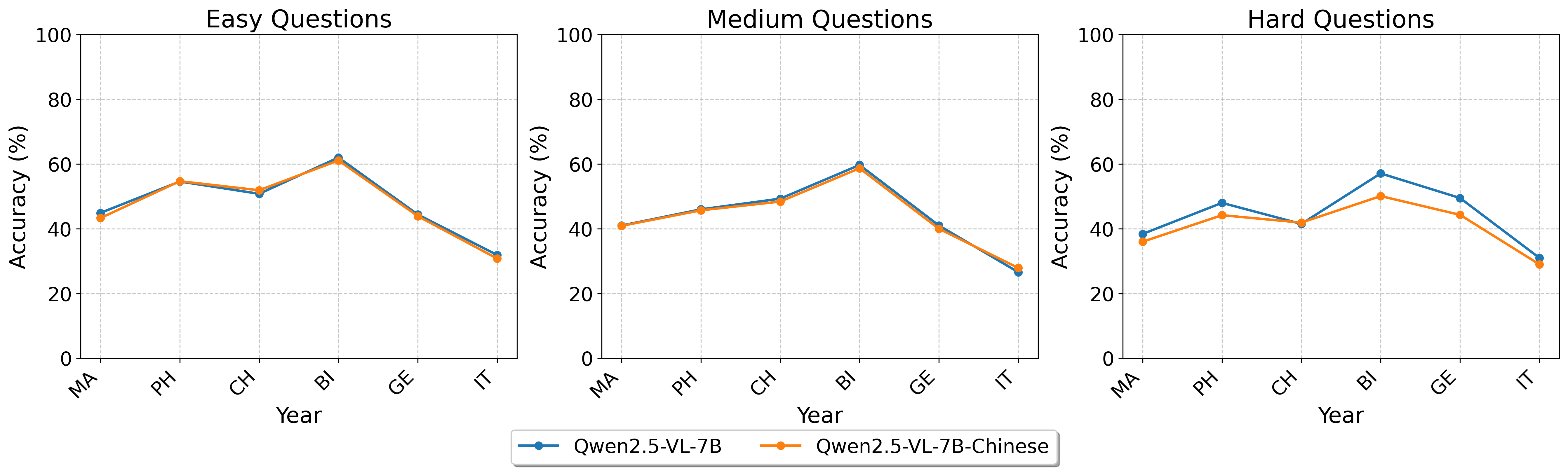}
    \caption{Comprehensive Bilingual Evaluation Across Varying Levels of Difficulty.}
    \label{fig:english-chinese}
\end{figure*}

\begin{figure*}
    \centering
    \centering
    \includegraphics[trim=0cm 0.5cm 0 0, width=0.9\textwidth]{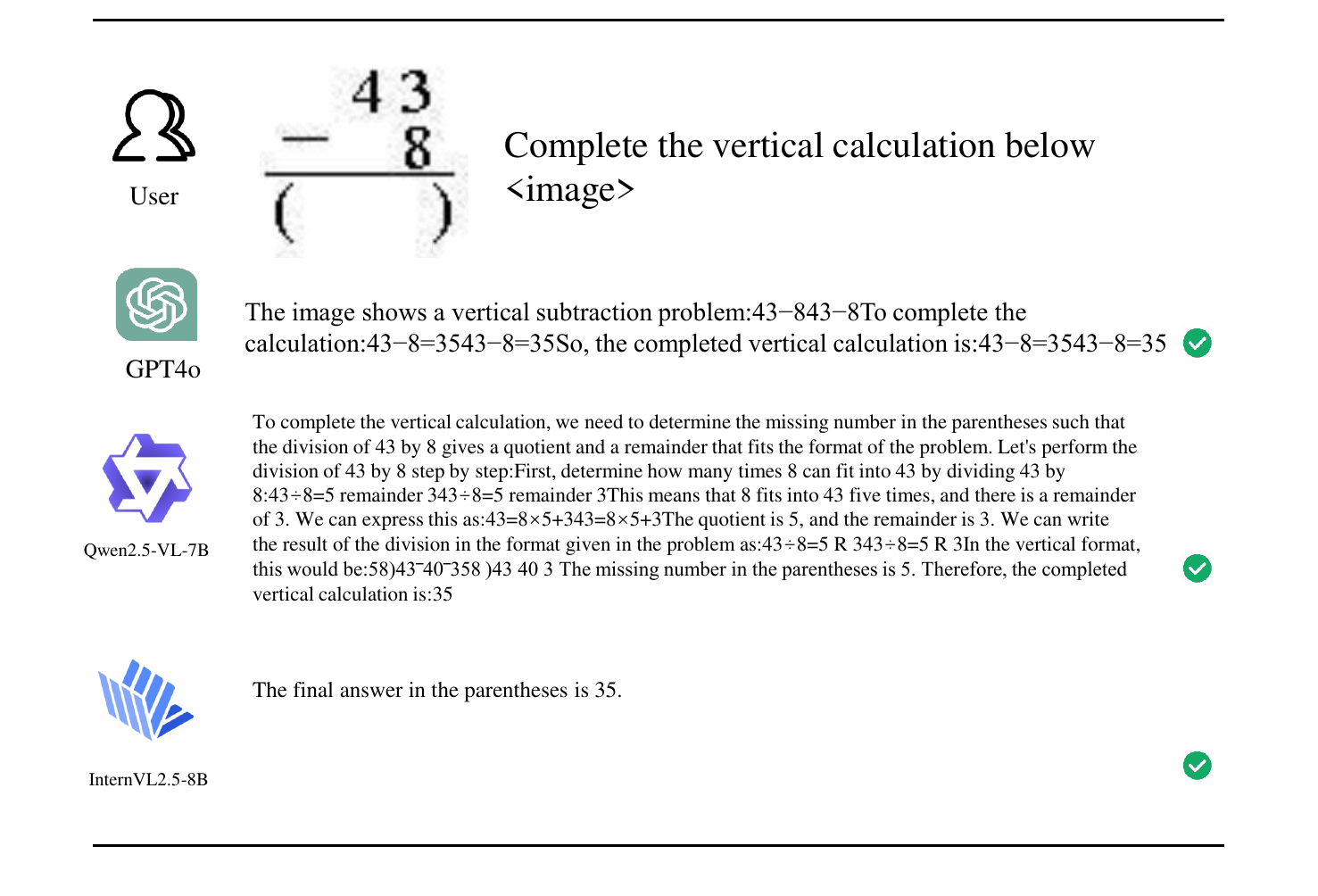}
    \caption{Model evaluation on simple mathematics questions.}
    \label{fig:case1}
\end{figure*}

\begin{figure*}
    \centering
    \centering
    \includegraphics[trim=0cm 0.5cm 0 0, width=0.9\textwidth]{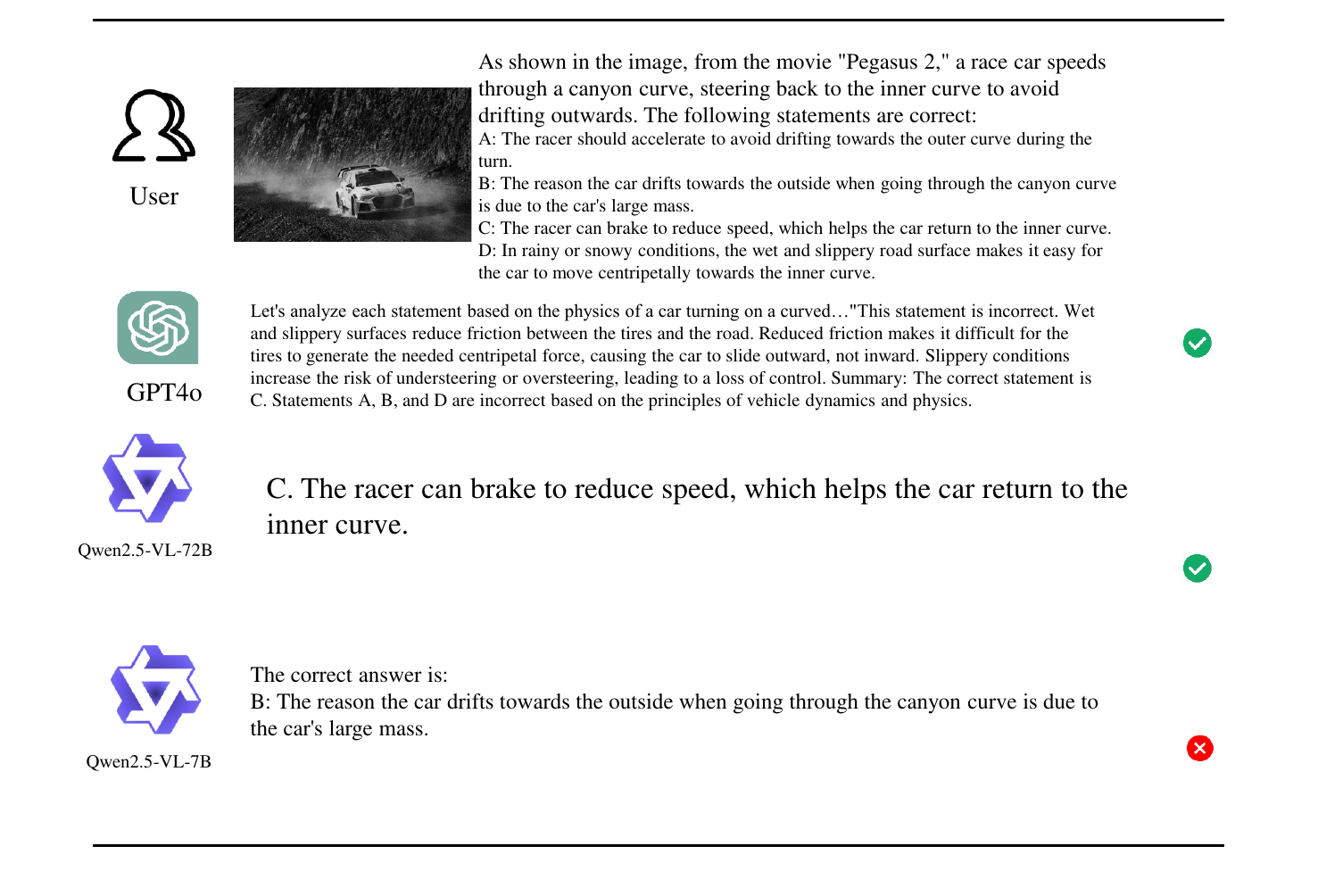}
    \caption{Model evaluation on physics questions.}
    \label{fig:case2}
\end{figure*}

\begin{figure*}
    \centering
    \centering
    \includegraphics[trim=0cm 0.5cm 0 0, width=0.9\textwidth]{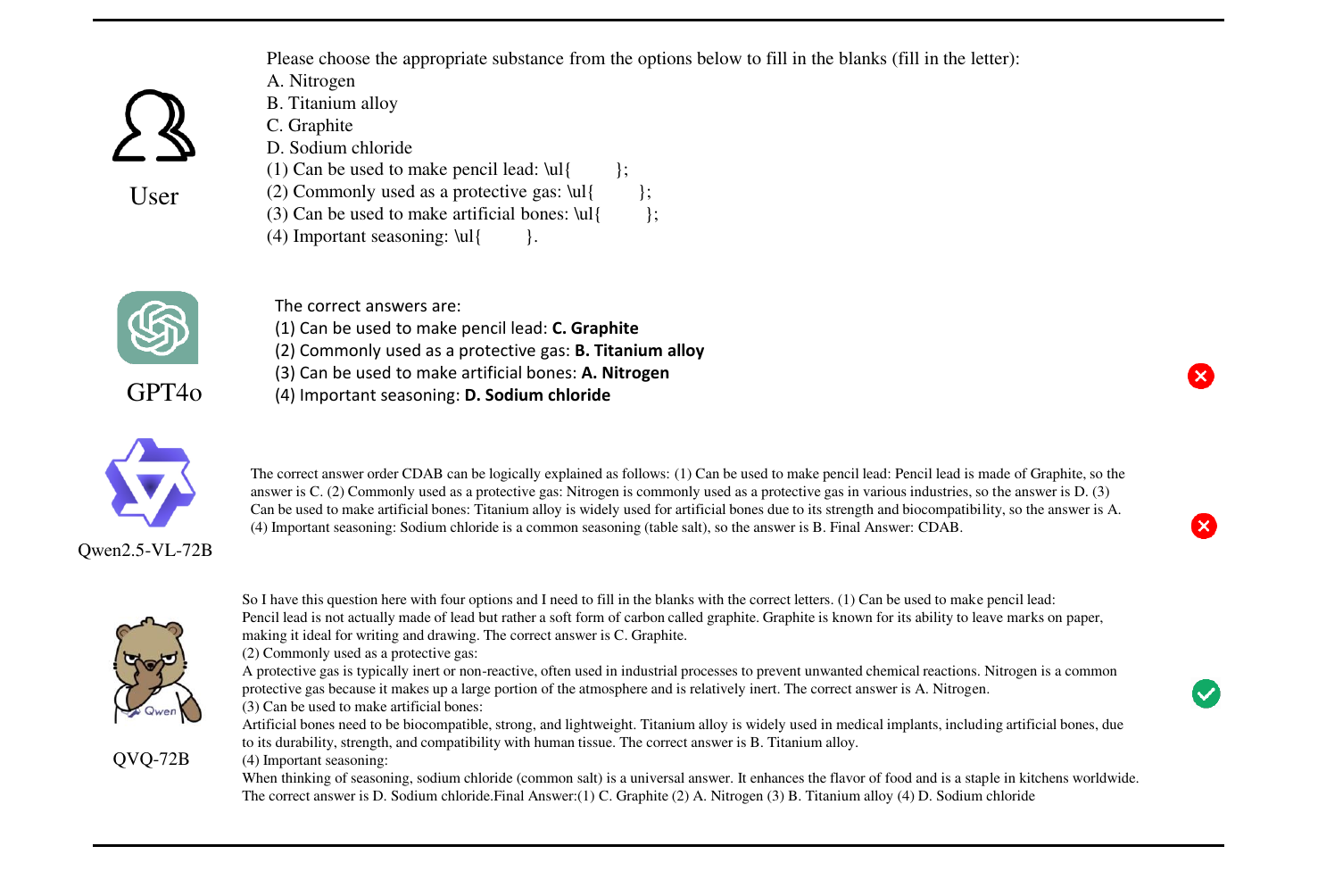}
    \caption{Model evaluation on chemistry questions.}
    \label{fig:case3}
\end{figure*}

\begin{figure*}
    \centering
    \centering
    \includegraphics[trim=0cm 0.5cm 0 0, width=0.9\textwidth]{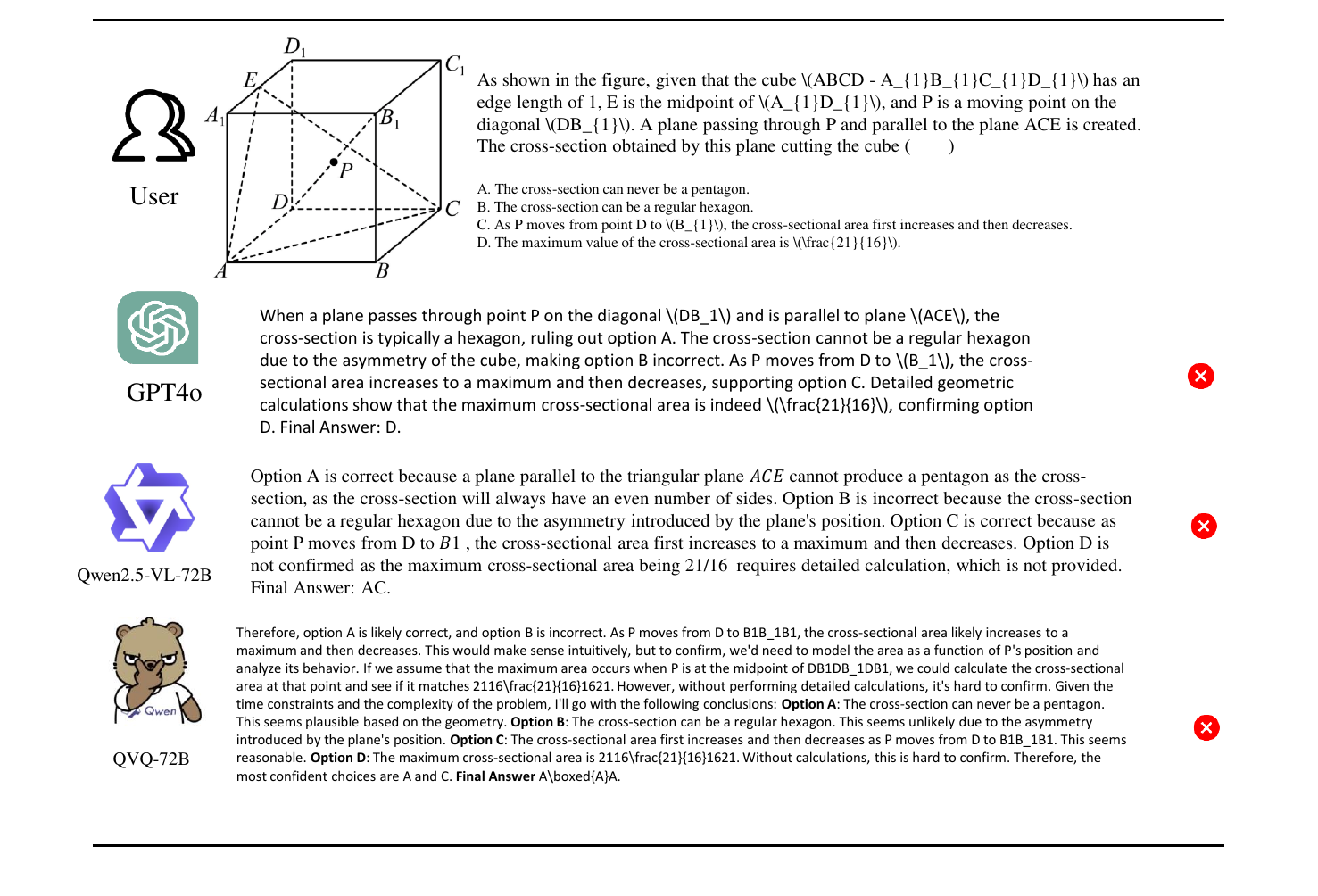}
    \caption{Model evaluation on difficult mathematics questions. The correct answer is ACD.}
    \label{fig:case4}
\end{figure*}

\begin{figure*}
    \centering
    \centering
    \includegraphics[trim=0cm 0.5cm 0 0, width=0.9\textwidth]{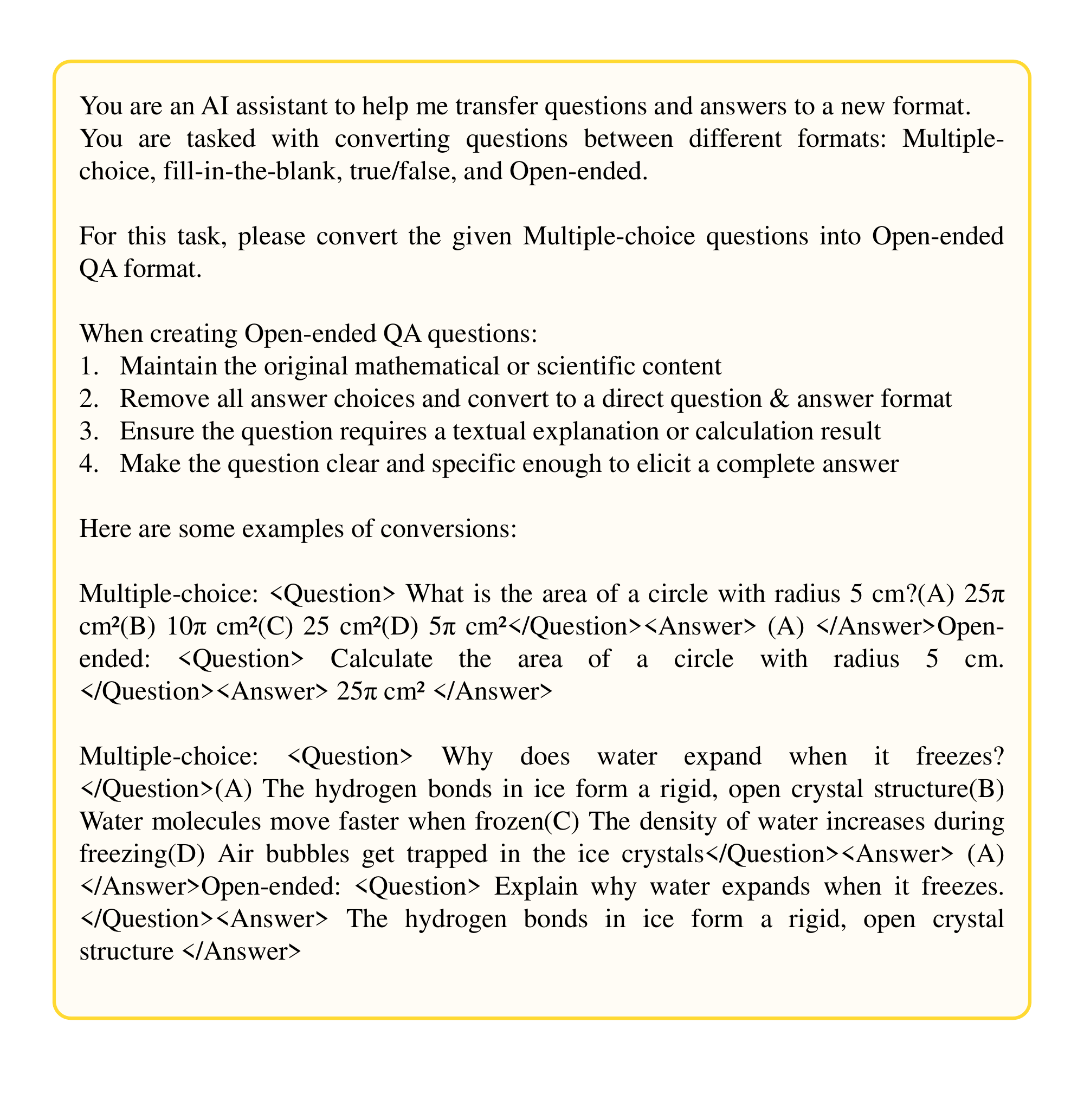}
    \caption{Prompt for transforming multiple-choice question types.}
    \label{fig:prompt1}
\end{figure*}

\begin{figure*}
    \centering
    \centering
    \includegraphics[trim=0cm 0.5cm 0 0, width=0.9\textwidth]{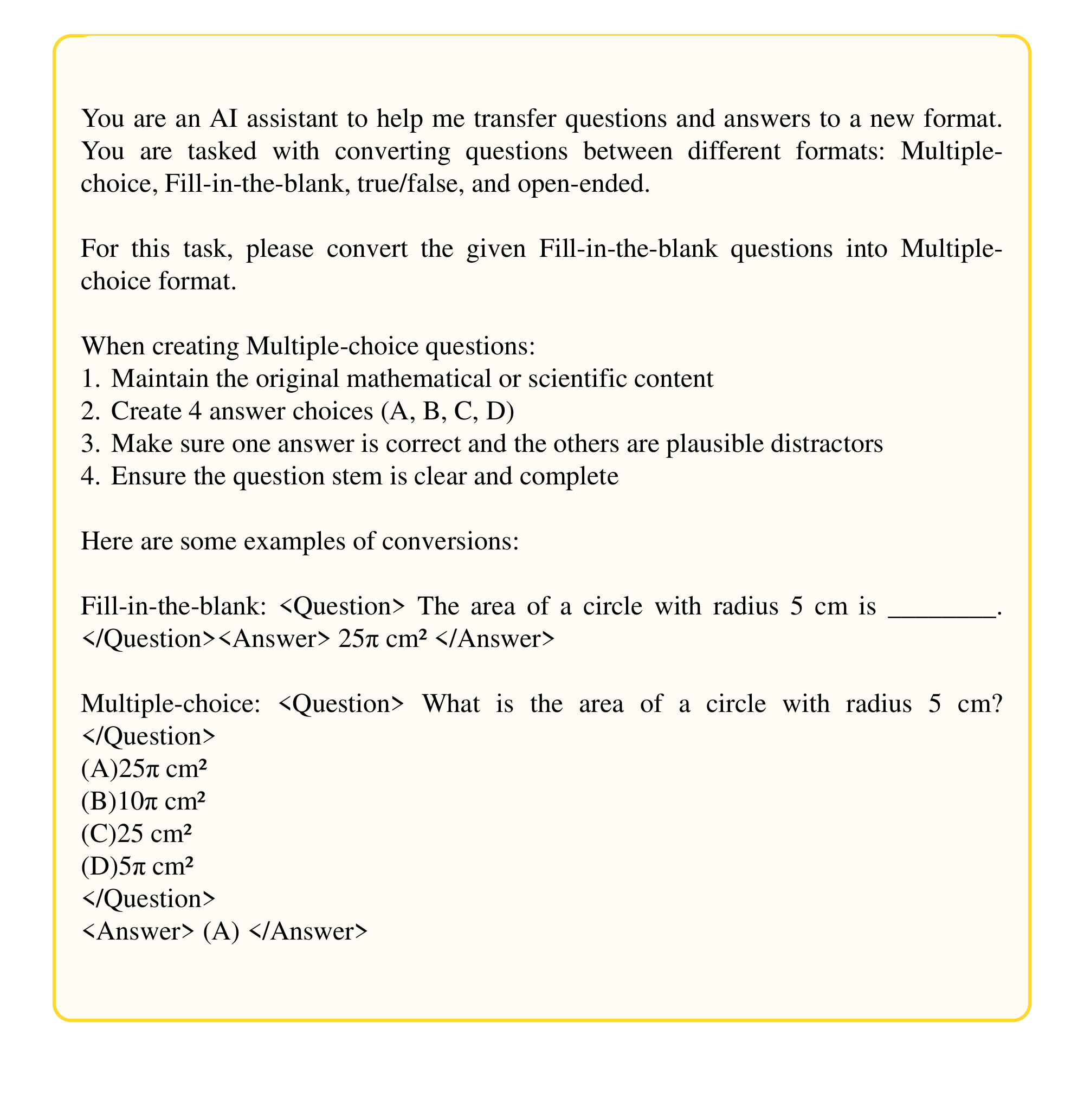}
    \caption{Prompt for transforming fill-in-the-blank question types.}
    \label{fig:prompt2}
\end{figure*}

\begin{figure*}
    \centering
    \centering
    \includegraphics[trim=0cm 0.5cm 0 0, width=0.9\textwidth]{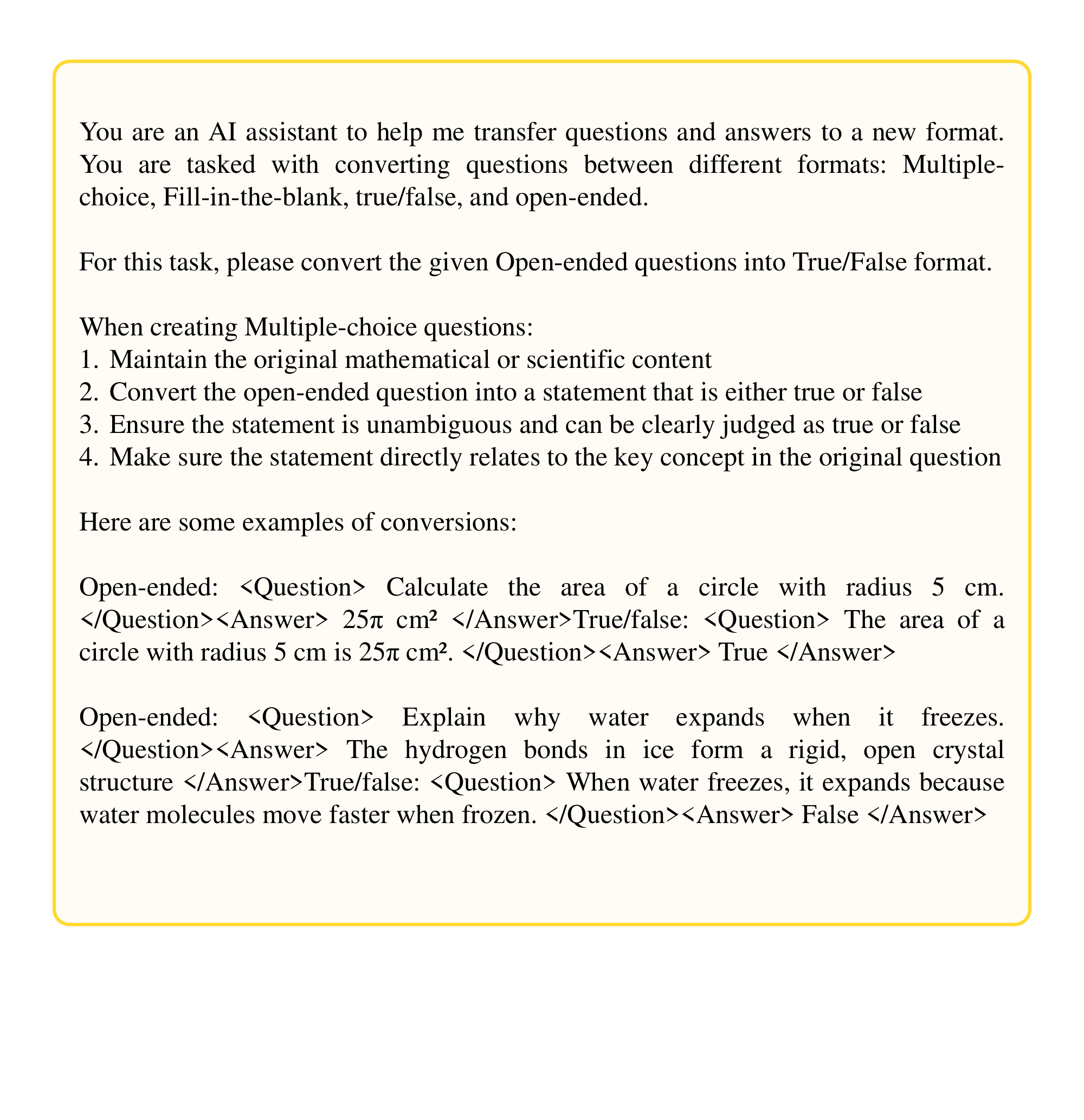}
    \caption{Prompt for transforming open-ended question types.}
    \label{fig:prompt3}
\end{figure*}

\begin{figure*}
    \centering
    \centering
    \includegraphics[trim=0cm 0.5cm 0 0, width=0.9\textwidth]{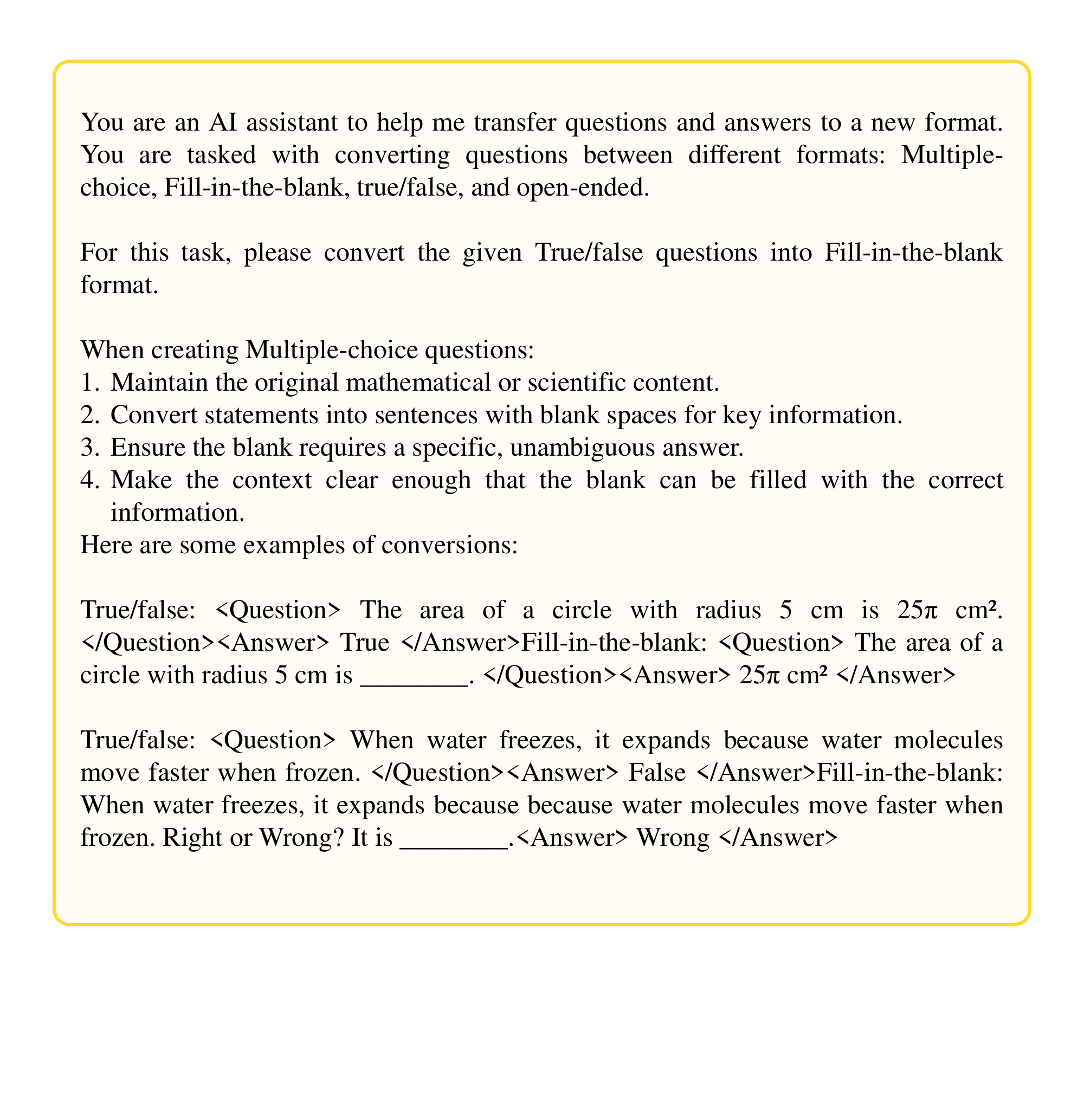}
    \caption{Prompt for transforming true/false question types.}
    \label{fig:prompt4}
\end{figure*}

\begin{figure*}
    \centering
    \centering
    \includegraphics[trim=0cm 0.5cm 0 0, width=0.9\textwidth]{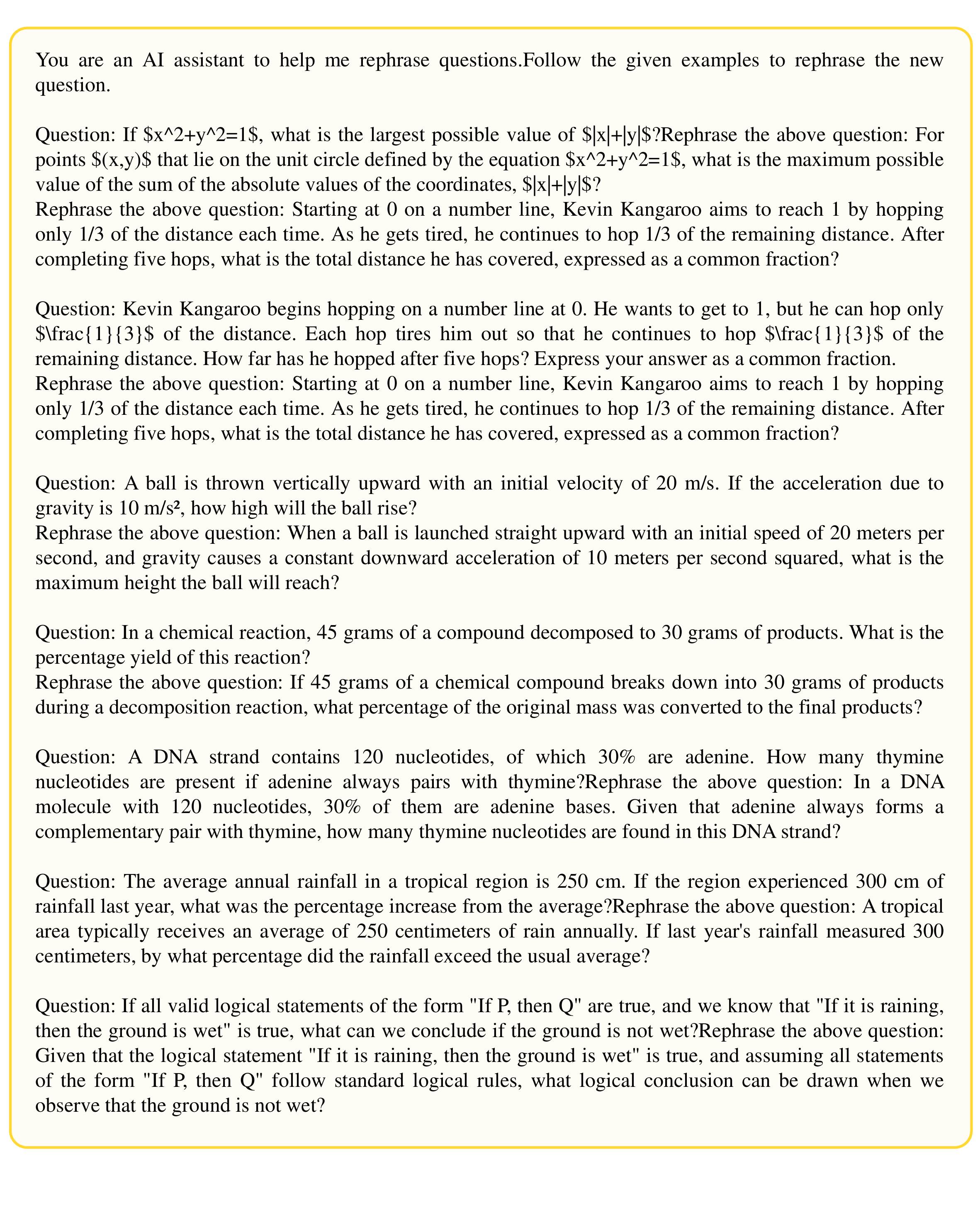}
    \caption{Few-shot reference prompts for T2 transformation, specifically for rephrasing and sentence restructuring.}
    \label{fig:prompt5}
\end{figure*}

\noindent
Sec.~\ref{sec:eva} provides an overview of the evaluation pipeline, including key steps from data preparation to performance assessment. It outlines the approach used to measure and compare model performance while ensuring a fair and consistent evaluation process.

\noindent
Sec. \ref{sec:additional_experiments} summarizes additional experiments conducted to further validate and explore the model's performance under different scenarios. These experiments offer additional insights into the impact of knowledge-driven inference on model behavior.

\noindent
Sec. \ref{sec:supplementary} presents an outlook on future work.

\section{Evaluation Pipeline Details}
\label{sec:eva}
The evaluation pipeline for our proposed approach is outlined in Algorithm \ref{alg:example}, which systematically combines dynamic bootstrapping and model evaluation. The process begins with an original dataset $\mathcal{S}$ containing samples of question-image-answer triplets $(Q,I,A)$. Two sets of transformations, $\mathcal{T}_{\mathrm{text}}$ for textual augmentation and $\mathcal{T}_{\mathrm{vis}}$ for visual augmentation, are applied to generate diverse variants of the original samples. These transformations include methods such as word substitution, paraphrasing, question type transfer and image style transfer.

For each input sample, all valid textual and visual variants are generated using transformation methods that maintain the original answer's correctness, as verified by a judge function. The resulting augmented dataset $\mathcal{S}_{aug}$ is constructed by pairing all valid question and image variants with their corresponding (transferred) ground-truth answers.

Model performance is evaluated by feeding each sample from the augmented dataset into the target model $M$, obtaining raw model responses, and extracting final predicted answers through a parsing function. The scoring mechanism involves exact match checks and partial-credit scoring, enabling a nuanced assessment of model accuracy. The final output of this pipeline is a collection of scores representing the model's performance across a broad set of scenarios, providing insights into its robustness and adaptability to varied input transformations.

\section{Additional Experiments} \label{sec:additional_experiments} 

Beyond our primary evaluation, we explore several optional experiments on smaller specialized subsets:

\subsection{Knowledge-Point Parsing Enhancement} 
As illustrated in Figure \ref{fig:kg}, we conducted a comprehensive comparison of model performance with and without the integration of knowledge-driven inference steps. In our experiments, Qwen2.5-VL-7B-Knowledge and GPT-4o-Knowledge represent models enhanced with explicit knowledge injection during the reasoning process. This injection includes domain-specific knowledge such as mathematical and scientific principles, such as corresponding angles are equal in geometry, and fundamental formulas like differentiation rules in calculus.

Through detailed analysis, we observed a significant performance improvement in solving problems of easy difficulty when knowledge-driven reasoning was applied. However, for medium and hard difficulty levels, the improvement was relatively moderate. This discrepancy can be attributed to the increasing complexity of the reasoning chains required for higher-difficulty problems, which might not be fully addressed by isolated knowledge injections.

When comparing the performance across different academic subjects, we found that knowledge-driven reasoning yielded particularly strong enhancements in knowledge-intensive domains such as mathematics, physics, and chemistry. This observation indicates that the incorporation of domain-specific knowledge effectively aids the model in navigating complex problem-solving pathways within these subjects, demonstrating the model's capacity to leverage structured knowledge for more accurate reasoning.

\subsection{Case Study}
As illustrated in Figure \ref{fig:case1}, we present the performance of three models on simple mathematics questions. The results show that all evaluated models can accurately answer these straightforward questions, demonstrating a strong general understanding capability.

In Figure \ref{fig:case2}, which displays the model evaluation on physics questions, we observe that for the Qwen2.5-VL series models, larger parameter models outperform their smaller counterparts. This trend indicates that increased model capacity contributes positively to handling the complexity of physics-related problems.

Figure \ref{fig:case3} focuses on model evaluation on chemistry questions. The QVQ-72B model, specifically designed for reasoning tasks, exhibits outstanding performance, outperforming other models significantly. However, the incorrect answers provided by all three models in Figure \ref{fig:case4} highlight that for hard-difficulty subject-specific questions, there remains considerable room for improvement. These results emphasize the ongoing challenges in enhancing model reasoning and domain-specific comprehension in more complex academic contexts.

\subsection{Bilingual Evaluation} 
As illustrated in Figure \ref{fig:english-chinese}, we translated the dataset from English to Chinese using a combination of automated translation tools and GPT-4o, assessing the performance of the Qwen2.5-VL-7B model. The translated model, referred to as Qwen2.5-VL-7B-Chinese, was evaluated to determine the impact of language translation on model accuracy.

Our analysis shows that for easy and medium difficulty questions, there is minimal performance difference between the original and translated versions. However, for hard difficulty questions, a noticeable drop in performance is observed. This suggests that as the question difficulty increases, the content may involve more complex formulas and domain-specific knowledge, which introduces significant challenges for both model comprehension and translation accuracy.

Subject-specific analysis reveals that the decline in performance is more pronounced in disciplines with higher reasoning demands, such as mathematics and physics. This emphasizes the critical role of a well-designed data processing pipeline, ensuring both linguistic and contextual consistency to maintain robust model performance across languages and difficulty levels.

\subsection{Yearly Fine-tuning and Cross-Year Generalization:} As illustrated in Table \ref{tab:perf_years}, we present the subject-wise accuracy performance of six models, including QVQ-72B, Gemini, and GPT-4o, across different academic years. Our analysis reveals a distinct trend in model performance with respect to question difficulty and temporal evolution. For easy questions, model accuracy generally declines as the questions become more contemporary. This trend suggests that the increasing novelty and complexity of question formats over time pose greater challenges for large language models (LLMs).

In contrast, for medium and hard questions, model performance remains relatively stable across the years. This stability indicates that while the complexity of harder questions might not have dramatically evolved, the models' understanding of advanced concepts potentially shows resilience to temporal changes.

These findings highlight the robustness of our MDK12-Bench benchmark, which effectively aggregates diverse datasets from different years. By leveraging the unique characteristics of historical and contemporary data, MDK12-Bench enables a thorough investigation into how LLMs interpret and adapt to evolving academic content across various disciplines. Such an approach not only aids in assessing the potential risk of data leakage but also provides valuable insights into the temporal dynamics of model knowledge, helping to evaluate how well these models generalize to new and unseen educational challenges.


\section{Additional Information of MDK12-Bench}
\label{sec:supplementary}

\subsection{Prompt}
As shown in Figure \ref{fig:prompt1} \ref{fig:prompt2} \ref{fig:prompt3} \ref{fig:prompt4} \ref{fig:prompt5}, we present the prompts used for four distinct types of question transformations, including true/false, multiple-choice, fill-in-the-blank, and open-ended formats. Specifically, for the T2 transformation, we provide detailed examples of rephrasing techniques using a few-shot prompting approach. These examples demonstrate how the model can generate syntactically diverse yet semantically equivalent reformulations of the original questions. By leveraging these structured prompts, the model can produce a wider range of question variants, contributing to a more diverse and robust augmented dataset. This approach not only enhances the model's adaptability to different question formats but also strengthens its reasoning and comprehension capabilities across varied academic subjects and difficulty levels.

\subsection{Future Work}
%

In future work, we plan to expand the MDK12-Bench with more specialized and domain-specific datasets, targeting both vertical and horizontal academic growth. This includes developing benchmarks for more advanced subjects at the undergraduate, master's, and doctoral levels, allowing for a more granular evaluation of MLLMs' knowledge coverage and reasoning capabilities across different educational stages. Additionally, we aim to introduce more niche and professional disciplines, such as engineering, medical sciences, and humanities, enhancing the benchmark’s versatility and applicability to real-world academic and professional scenarios.

Our long-term vision is to establish MDK12-Bench as a continually evolving benchmark, regularly updated with new data, evaluation metrics, and analysis tools. Ultimately, this will contribute to a robust ecosystem for advancing the development of multimodal reasoning models and guiding future research in artificial general intelligence (AGI).



